**Development and external validation of a multimodal artificial intelligence mortality prediction model of critically ill patients using multicenter data**

Behrooz Mamandipoor, BS[1,2]; Chun-Nan Hsu, PhD[2]; Martin Krause, MD[2,3]; Ulrich H. Schmidt, MD, PhD, MBA[3]; Rodney A. Gabriel, MD, MAS[1,2]

[1]Department of Biomedical Informatics, University of California, San Diego Health, La Jolla, CA, USA

[2]Division of Perioperative Informatics, Department of Anesthesiology, University of California, San Diego, La Jolla, CA, USA

[3]Division of Critical Care Medicine, Department of Anesthesiology, University of California, San Diego, La Jolla, CA, USA

**Corresponding author:** Rodney A. Gabriel, Email: ragabriel@health.ucsd.edu

**Previous presentations:** none

**Acknowledgements:**

**Word count:**

Abstract: 299

Introduction: 452

Discussion: 856

Figures: 5

Tables: 2

Supplemental Files: 19

**Abbreviated Title:** Multimodal deep learning for predicting mortality

**Summary Statement:** Our findings highlight the importance of incorporating multiple sources of patient information for mortality prediction and the importance of external validation. Our data harmonization leads to robust models for mortality prediction.

**Funding Statement:** This study is funded by the National Institutes of Health 1OT2OD037995-01

**Conflicts of Interest:** Dr. Gabriel's institution, the University of California, has received funding and/or product for research from Wellcome Leap (Los Angeles, CA), Advanced Research Projects Agency for Health (Bethesda, MD), National Institutes of Health (Bethesda, MD), Anesthesia Patient Safety Foundation (Schaumburg, Illinois), Avanos Medical (Alpharetta, GA), Pacira Biosciences (Tampa, FL), Takeda (San Diego, CA), and Merck (Rahway, NJ).

**Author Contributions:** BM was responsible for study design, data collection, data pre-processing, analysis, figure/table design, and manuscript preparation. CH was responsible for study design, analysis, figure/table design, and manuscript preparation. MK was responsible for study design, figure/table design, and manuscript preparation. US was responsible for study design, figure/table design, and manuscript preparation. RG was responsible for study design, data collection, data pre-processing, analysis, figure/table design, and manuscript preparation

**Competing Interest Statement:** none

**Classification:** Artificial Intelligence, critical care medicine, predictive modeling

**Abstract**

**Background:** Early prediction of in-hospital mortality in critically ill patients can aid clinicians in optimizing treatment. The objective was to develop a multimodal deep learning model, using structured and unstructured clinical data, to predict in-hospital mortality risk among critically ill patients after their initial 24 hour intensive care unit (ICU) admission.

**Methods:** We used data from MIMIC-III, MIMIC-IV, eICU, and HiRID. A multimodal model was developed on the MIMIC datasets, featuring time series components occurring within the first 24 hours of ICU admission and predicting risk of subsequent inpatient mortality. Inputs included time-invariant variables, time-variant variables, clinical notes, and chest X-ray images. External validation occurred in a temporally separated MIMIC population, HiRID, and eICU datasets. Area under the receiver operating characteristics (AUROC), area under the precision-recall curves (AUPRC), and Brier scores were reported.

**Results:** A total of 203,434 ICU admissions from more than 200 hospitals between 2001 to 2022 were included, in which mortality rate ranged from 5.2% to 7.9% across the four datasets. The model integrating structured data points had AUROC, AUPRC, and Brier scores of 0.92 [95% confidence interval (CI) 0.90 – 0.93], 0.53 [95% CI 0.49 – 0.57], and 0.19 [95% CI 0.18 – 0.20], respectively. We externally validated the model on eight different institutions within the eICU dataset, demonstrating AUROCs ranging from 0.84-0.92. When including only patients with available clinical notes and imaging data, inclusion of notes and imaging into the model, the AUROC, AUPRC, and Brier score improved from 0.87 [95% CI 0.85 – 0.89] to 0.89 [95% CI 0.87

– 0.91], 0.43 [95% CI 0.36 – 0.51] to 0.48 [95% CI 0.40 – 0.56], and 0.37 [95% CI 0.36 – 0.39] to 0.17 [95% CI 0.16 – 0.19], respectively.

**Conclusions:** Our findings highlight the importance of incorporating multiple sources of patient information for mortality prediction and the importance of external validation.

**Introduction**

Mortality is an important clinical outcome in patients admitted to the intensive care unit (ICU).[1] In the first few hours of ICU admission, decisions about management and interventions impact the patient's chance of survival.[2] Timely prediction of in-hospital mortality in critically ill patients can aid in benchmarking of ICU outcomes, triggering of evaluation for palliative care or withdrawal of care early on, or as an early warning system to better evaluate patients for interventions that may be missing that would be effective in mitigating mortality.

Prior tools for predicting in-hospital mortality, such as the Simplified Acute Physiology Score II,[3] Acute Physiology and Chronic Health Evaluation IV,[4] Sequential Organ Failure Assessment,[5] and Oxford Acute Severity of Illness Score[6] relied on time-invariant data, including demographic information, laboratory results, and vital signs. These models have over-predicted mortality when applied to more recent datasets, requiring frequent adjustment.[7] In recent years, electronic health records (EHR) have made collecting vast amounts of clinical data easier, which enabled advanced machine learning models to predict patient outcomes in healthcare.[8] However, many deep learning predictive models of ICU outcomes typically used EHR-derived clinical data from a single modality.[9] Incorporating unstructured clinical data modalities, including clinical text notes and medical imaging,[10-11] may improve predictive performance.[12-14]

The objective of this study was to develop an artificial intelligence (AI) model using multimodal data that predicts risk of mortality among critically ill patients (defined as inpatient death occurring any time after the initial 24 hours of an ICU admission) and validate the model's generalizability on separate datasets. A key issue of any AI predictive model in clinical workflow is trustworthiness. Clinicians will be concerned about false alerts, explainability, generalizability to diverse populations, and fairness/bias towards under-represented social cohorts.[15] Thus, in addition to demonstrating generalizability, we aim to provide transparency of any biases from our model's predictions towards specific demographic groups based on age, sex, race, and ethnicity.

**Materials and Methods**

The study utilized de-identified publicly available data from Medical Information Mart for Intensive Care (MIMIC)-III, MIMIC-IV, eICU Collaborative Research Database, and high time-solution ICU dataset (HiRID) and was therefore exempt from the patient informed consent requirement by the institutional review board (Human Research Protections Program). The objective of the study was to develop multimodal deep learning models (incorporating time-invariant and time-variant data including vital signs, medication administration, ventilator changes, laboratory values, validated risk scores, chest X-ray images, and clinical notes) for predicting inpatient mortality (at any time point during their hospital stay) after the initial 24 hours a patient is in the intensive care unit. This was accomplished by using MIMIC datasets as development and internal validation sets and using HiRID and eICU as external validation sets. Data analysis planning was performed after accessing the datasets.

*Description of Datasets*

We utilized the following EHR datasets: MIMIC-III,[16-17] MIMIC-IV,[18-19] eICU,[20-21] and HiRID.[22-23] The MIMIC-III database is a widely accessible critical care database that integrates de-identified clinical data from patients admitted to intensive care units at the Beth Israel Deaconess Medical Center in Boston, Massachusetts. This database encompasses records from 53,423 hospital admissions from 2001 to 2012. MIMIC-IV extends this resource by including clinical data from more than 90,000 ICU admissions from 2008 to 2022 at the same institution. This dataset includes details on each patient's hospital stay, encompassing laboratory measurements, medications, vital signs, orders, diagnoses, procedures, treatments, and de-identified free-text clinical notes and chest

X-ray images. The eICU Collaborative Research Database represents a multi-center database specifically tailored to meet the need of research for intensive care units, containing detailed data with high granularity from over 200,859 ICU admissions, pertaining to 139,367 distinct patients admitted across 335 units in 208 hospitals between 2014 and 2015 across the United States. The HiRID database is a publicly accessible dataset that captures critical care data from more than 33,000 patients admitted to the Department of Intensive Care Medicine at the University Hospital of Bern, Switzerland, during the period from January 2008 to June 2016. Neither eICU nor HiRID contained clinical notes and chest X-rays. Thus, only models integrated time variant and -invariant clinical data were externally validated. Patient inclusion and exclusion for eICU (**Supplement Figure 1a**), HiRID (**Supplement Figure 1b**), MIMIC-III (**Supplement Figure 1c**), and MIMIC-IV (**Supplement Figure 1d**) are provided in the supplementary materials. Detailed distribution of data points for eICU, HiRID, MIMIC-III, and MIMIC-IV is provided in **Supplement Tables 1-4**, respectively.

*Study Design*

The primary objective of our study was to develop a multimodal deep learning model incorporating time-variant and time-invariant clinical data, imaging, and clinical notes to predict inpatient mortality after the initial 24 hours in the ICU (outcome is defined as death at any time in the hospital after this 24 hour initial time period). All data incorporated into the model were only those known during that initial 24 hour time period. As our model was designed to predict mortality after 24 hours of ICU admission, we excluded ICU stays shorter than 24 hours. We leveraged five data types within the datasets: 1) time-invariant structured data – defined as static structured data points that did not change with time. These included baseline comorbidities and demographic data (e.g., race/ethnicity, sex, age, body mass index); 2) dynamic structured data points such as laboratory values, administered medications, calculated clinical scores (e.g., risk scores, Glasgow Coma Scale), ventilator and oxygen settings, fluid intake/output, and procedures/interventions; 3) vital signs with dynamic measurements such as heart rate, respiratory rate, and oxygen saturation; 4) clinical notes; and 5) chest x-ray images. Data types 2 and 3 are time-variant clinical data – defined as dynamic structured data points that changed with time and were further divided into two sub-categories. A detailed description of the variable definitions are provided in **Supplement Text 1**. Several data harmonization and normalization steps were conducted for each data type prior to model development (**Supplement Figure 1**). Continuous variables were normalized to have a zero mean and unit variance. Data harmonization code may be found at https://github.com/bmamandipoor/ICU-Mortality/tree/main. We then developed various versions of our predictive model using combinations of each data type. The model was trained via multimodal deep learning, in which time-invariant data was processed via a multilayer perceptron neural network, time-variant data via a bidirectional long short-term memory network, clinical notes via a pretrained transformer language model (BERT), and chest X-ray images via a convolutional neural network. Embedding representations of the output of each component were then concatenated and processed via a pooling layer prior to integration into a classification head for final calculated probability of the outcome (**Figure 1**). Models were both internally and externally validated to measure generalizability.

To ensure fairness/bias of our models across various demographic cohorts (based on race, ethnicity, age, and sex), we conducted a bias audit using the Aequitas toolkit - an open-source library for bias and fairness analysis.[24] We calculated the false positive, false negative, true positive, and true negative rates of the model predictions for each subset and plotted the rates. For age, we compared these rates based on age under 45, between 45 – 60, between 61 – 75, and over 76 years old.

We also reported examples of model explainability on our validation sets. We adopted the Integrated Gradients (IG) method[25] to address the interpretability of our long short-term memory and BERT models used in analyzing dynamic EHR data and clinical notes, respectively. For clinical notes, we generated our interpretation of the BERT model using the LayerIntegratedGradients method from the Captum library.[26] For interpretation of the DenseNet121 model used to analyze chest x-ray data, we applied Gradient-weighted Class Activation Mapping (Grad-CAM).[27-28] Detailed description of the model evaluation is provided in **Supplement Text 2**.

We sought to assess the individual contribution of each data modality – time-invariant data, time-variant data, vital signs, clinical notes, and imaging – and their combinations based on modality availability across patient subgroups. While vital signs are a class of time-variant data, we separated it out as its own data modality due to their continuous nature and high temporal resolution characteristics. Since all patients had time-variant data recorded, we used this modality as the baseline and incrementally incorporated additional modalities to assess their additive predictive value in four steps: *First,* for patients with time-invariant data, we evaluated models using this modality alone and in combination with the hourly measured vital signs and time-variant features (**Figure 2A**). *Second*, in a subgroup of patients who had clinical notes, we evaluated models using the clinical notes alone and integrated them with other structured data modalities (**Figure 2B**). *Third*, in a smaller subgroup of patients with chest x-rays, we assessed models using imaging alone and within a comprehensive multimodal model incorporating all available modalities (**Figure 2C**).

*Multimodal Model Architecture*

The architecture of our multimodal deep learning model consists of four types of neural networks (**Figure 2**):

**Multilayer Perceptron for time-invariant variables.** The multilayer perceptron consisted of two layers with 32 nodes for the first layer and 16 nodes for the second layer, each employing sigmoid activation functions. Batch normalization and dropout layers were interleaved between the layers.

**Bidirectional Long Short-Term Memory for time-variant variables.** The long short-term memory network had hidden sizes of 128. The input sequences from both directions were concatenated and passed through a classification head consisting of a linear layer with 32 units, a rectified linear unit activation function, and a dropout layer.

**BERT Language Model for clinical notes.** The model utilized the pre-trained BERT (Bidirectional Encoder Representations from Transformers) architecture.[29] Since each patient may had multiple clinical notes, the model processed each note individually through the BERT model to obtain its corresponding pooler output. We then gathered all the pooler outputs for each patient and passed them through a long short-term memory module with 128 units. The final representation from the long short-term memory model was then passed through a classification head composed of two linear layers with 128 and 32 units, respectively. These layers incorporated gaussian error linear unit activation functions and dropout layers.

**DenseNet Convolutional Neural Network for chest X-rays.** The model leveraged the pre-trained DenseNet121 convolutional neural network architecture[30] to process and extract visual embeddings from each patient's chest X-rays imaging data. Similar to clinical notes, since each patient may have had multiple chest X-ray images, the model processed each image individually through the DenseNet121 model to obtain its corresponding visual embedding. We then gathered all the visual embeddings for each patient and passed them through a long short-term memory module with 128 units. The final representation from the long short-term memory was then passed through a classification head comprising two linear layers with 128 and 32 units, respectively. These layers incorporated gaussian error linear unit activation functions and dropout layers.

**Pooling Layer and Classification Head.** The pooling layer aggregated output embeddings from each deep learning model into fixed-length embeddings using one of the following methods: 1) concatenation, 2) element-wise addition, 3) self-attention,[31] or 4) deep & cross network.[32-33] Finally, the fixed-length embeddings coming from the pooling layer subsequently passed through the final classification head, which featured two linear layers with 128 and 32 units, respectively, and incorporated rectified linear unit activation functions.

*Multimodal Model Development*

On the development sets (MIMIC datasets), data was split randomly into a 70% training, 10% validation (for hyperparameter tuning) and 20% test set (for internal validation). We trained our multimodal deep learning models by training the multilayer perceptron and long short-term memory models from scratch and fine-tuning pretrained BERT (clinical text) and DenseNet121 (chest X-ray images) models using our training data. After training the single-modality models, we fine-tuned the pooling layers and classification heads of the multimodal models, adjusting them based on the availability of data modalities for each patient group. During this fine-tuning process of the classification head, the single modality models remained frozen (their weights were not updated). This approach was chosen because end-to-end fine-tuning is resource-intensive and may not always lead to improved model generalization performance. To stabilize performance and facilitate convergence, we applied Xavier uniform initialization for the weights in linear layers and recurrent neural network components.[34] To address class imbalance, we employed weighted random sampling and a weighted cross-entropy loss function. By calculating class frequencies, we assigned inverse-proportional weights to create sampling probabilities that compensate for imbalance, ensuring that each training batch fairly represented both classes. The loss function weights prioritized the minority class, maintaining a balanced focus during training. For optimizing the multilayer perceptron and long short-term memory models, we used the Adam optimizer with an initial learning rate of 5e-4 and a weight decay of 0.01,[35] incorporating a StepLR scheduler to reduce the learning rate by a factor of ten every ten epochs. For the BERT and DenseNet121 models, we employed the AdamW optimizer.[36] We also employed differential learning rates[37] and the gradual unfreezing techniques.[38] We set the batch size to 16, halted training if no improvement in area under the receiver operating characteristics curve (AUROC) or F1 score was observed on the validation set over 20 epochs, and used a checkpoint system to save the model state when improvements occurred.

*Statistical Analysis*

Absolute standardized mean difference was calculated to measure baseline differences between cohorts for each dataset. We assessed model performance based on AUROC, area under the precision-recall curve (AUPRC), Brier score, F1-score, Matthews correlation coefficient, accuracy, and precision. To quantify uncertainty, we calculated 95% confidence intervals (CI) using the pivot bootstrap method with 1,000 resamples of test sets. AUROCs were compared using DeLong's test, with the significance level set to 0.05.

**Results**

*Study population*

We used four different ICU datasets representing various patient populations from different geographic regions: MIMIC-III, MIMIC-IV, eICU, and HiRID (**Table 1**). We included 46,983 patients from MIMIC-III, in which 3,406 (7.2%) had an inpatient death after the initial 24 hours. The median [quartiles] time to death from ICU admission was 114.0 [52.3, 235.2] hours. For MIMIC-IV, we included 60,087 patients, in which 3,788 (6.3%) had an inpatient death after the initial 24 hours. The median [quartiles] time to death from ICU admission was 98.5 [47.3, 206.4] hours. In the HiRID dataset, there were 16,642 patients, in which 1,323 (7.9%) had an inpatient death occurring sometime after the initial 24 hours of ICU stay. The median [quartiles] time to death from ICU admission was 62.4 [38.4, 122.4] hours. The eICU dataset consisted of 132,781 patients, in which 6,921 (5.2%) had an inpatient death occurring sometime after the initial 24 hours of ICU stay. The median [quartiles] time to death from ICU admission was 62.4 [38.4, 122.4] hours. **Supplement Figure 2** illustrates the distribution of time until mortality across each dataset. Using this data, we developed multimodal deep learning models for predicting inpatient mortality after the initial 24 hours in the ICU (all input data were only those known within that first 24 hours).

*Predicting mortality using time-variant and time-invariant structured clinical data*

First, we developed and validated a deep learning model using MIMIC-III and MIMIC-IV (prior to 2019) to assess performance of prediction with the combination of time-variant (e.g., vital signs, medications, laboratory values, ventilator settings, etc) and time-invariant variables (e.g., baseline comorbidities, demographics) on predictive performance (**Figure 2A** and **Table 2**). For MIMIC-III data, models using only time-variant variables achieved AUROC and AUPRC at 0.91 [95% CI 0.90 - 0.92] and 0.47 [95% CI 0.43-0.51], respectively (**Table 2** and **Figure 3a**). This is compared to worse performance when using only time-invariant variables, with AUROC and AUPRC at 0.72 [95% CI 0.70 - 0.74] and 0.18 [95% CI 0.16-0.20], respectively. Combining both data types improved AUROC to 0.92 [95% CI 0.90-0.93] and AUPRC to 0.53 [95% CI 0.49-0.57]. Similar performance was observed when the model was trained and internally validated on MIMIC-IV (**Figure 3b**). Bias was observed towards patients older than 75 years old (with higher false positive rates) when tested on both the MIMIC-III and IV population, and White population (higher false negative rates) when tested on MIMIC-IV population (**Supplement Figure 5**). Results in all performance metrics are provided in **Supplement Table 5**.

To evaluate the model's predictive performance over time, we calculated performance metrics across three time intervals following the initial 24 hours of observation using the MIMIC-IV database. In the first interval, positive cases were defined as deaths occurring within the first day after the initial 24-hour observation period, compared to all surviving patients, yielding an AUROC of 0.96 [95% CI 0.95 – 0.97] and AUPRC of 0.52 [95% CI 0.45 – 0.60]. In the second interval, positive cases were defined as deaths occurring between the second day and the end of the first week, with an AUROC of 0.91 [95% CI 0.90 – 0.92] and AUPRC of 0.31 [95% CI 0.26 – 0.36]. Finally, for the third interval, positive cases were defined as deaths occurring after the first week, resulting in an AUROC of 0.88 [95% CI 0.87 – 0.90] and AUPRC of 0.15 [95% CI 0.12 – 0.19]. Similar results were calculated when performed on MIMIC-III data.

Next, we determined if model performance changed when patients designated as transitioning to hospice at the end of their hospital stay were counted towards mortality. MIMIC-III and MIMIC-IV included data for hospice disposition. In our test cohorts, 0.2% (MIMIC-III) and 2.0% (MIMIC-IV) of admissions were labeled as discharged to hospice. To ensure robustness for analyses that extend beyond ICU discharge, we added sensitivity analyses on the MIMIC cohorts: (1) excluding hospice discharges, and (2) treating hospice as a composite adverse outcome (death or hospice). These sensitivity analyses did not change predictive performance of the models likely due to the limited number of patients labeled as discharged to hospice.

*External validation*

Next, we performed external validation of the model to assess its generalizability on three separate external validation sets, eICU, HiRID, and MIMIC-IV data during the COVID-19 pandemic era 2020-2022. For patients admitted during the COVID-19 period, the model achieved an AUROC of 0.92 [95% CI 0.91 – 0.93] and an AUPRC of 0.54 [95% CI 0.50 – 0.58] (**Figure 4a**). When tested on the HiRID dataset, the AUROC and AUPRC were 0.89 [95% CI 0.88 – 0.90] and 0.48 [95% CI 0.45 – 0.51], respectively (**Figure 4a**) with minimal algorithmic bias (**Supplement Figure 6a**). When tested on the entire eICU data, the AUROC and AUPRC were 0.87 [95% CI 0.86 – 0.87] and 0.34 [95% CI 0.33 – 0.36], respectively (**Figure 4a**). Next, we separately tested the model on eight different institutions within the eICU dataset, demonstrating AUROC ranging from 0.84-0.92 (**Figure 4b**). Performance metrics are provided in **Supplement Table 6**. There was no notable bias towards cohorts based on race, ethnicity, or sex (**Supplement Figure 6b**). Variable importance analysis on the test sets for HiRID and eICU was performed, which demonstrated the most impactful variables for the prediction (**Supplement Figure 7a and 7b**, respectively).

*Incorporation of clinical notes for predicting mortality*

We utilized MIMIC datasets to assess the combination of time-variant and time-invariant variables with clinical notes (**Figure 2B**). We excluded patients that did not have clinical notes data, in which the final sample size for MIMIC-III and MIMIC-IV were 44,074 and 46,765, respectively. With MIMIC-III data, a model incorporating only clinical notes achieved an AUROC and AUPRC at 0.86 [95% CI 0.843-0.873] and 0.39 [95% CI 0.350-0.430], respectively (**Table 2** and **Supplement Figure 8a**). When combined with time-variant and -invariant data, the AUROC and AUPRC were 0.91 [95% CI 0.90-0.92] and 0.51 [95% CI 0.47-0.55] (differences in AUROC, $p < 0.05$), respectively. Lower false positive rates were observed in the Hispanic population (**Supplement Figure 8b**). Regarding model performance, similar patterns in metrics were seen with MIMIC-IV (**Supplement Figure 8c**). Higher false negative rates were observed in the Hispanic population (**Supplement Figure 8d**). Results in all performance metrics are provided in **Supplement Table 7**. Explanatory models for clinical notes were executed on randomly chosen example patients. **Supplement Figure 9** illustrates an example of identified pertinent text related to mortality and survival.

Next, we sought to compare model performance when patients with clinical notes (present during the first 24 hours) that indicated transitions to withdrawing care were removed. Using the MIMIC-III dataset, we filtered all clinical notes within the first 24 hours using targeted phrases (e.g., withdrawal of care, withhold life support, comfort care, comfort measures only, palliative, end-of-life, code status changed, hospice). This audit identified 1.6% of notes corresponding to 1.5% of patients with withdrawal-of-care language. We re-evaluated model performance after excluding those patients from the validation dataset; the metrics remained unchanged, perhaps due to the very limited prevalence of such documentation within the initial 24-hour window.

*Incorporation of medical imaging for predicting mortality*

We utilized the MIMIC-IV dataset to assess performance of models when all data modalities, including chest X-ray images, were combined into a multimodal deep learning model (**Figure 2C**). We excluded patients that did not have clinical notes data and imaging data, in which the final sample size was 9,881. The AUROC for CXR images alone and all modalities combined were 0.76 [95% CI 0.721-0.794] and 0.89 [95% CI 0.866-0.913] (differences in AUROC, $p < 0.05$), respectively (**Table 2** and **Supplement Figure 10a**). The AUPRC for CXR images alone and all modalities combined were 0.25 [95% CI 0.196-0.313] and 0.48 [95% CI 0.398-0.555], respectively. There were decreased false negative rates in the Black population and female sex (**Supplement Figure 10b**). Results in all performance metrics are provided in **Supplement Table 8**. Explanatory models for CXR images were executed on a randomly chosen deceased patient. **Supplement Figure 11** illustrates an example of an identified CXR image related to mortality.

*Comparison to current validated risk scores*

We calculated four widely recognized clinical severity scores and mortality estimation systems. These include the Sequential Organ Failure Assessment, Simplified Acute Physiology Score II, Oxford Acute Severity of Illness Score, and the Acute Physiology and Chronic Health Evaluation II. Using MIMIC-III data, out of the current risk scores, Simplified Acute Physiology Score II performed the best among all risk scores considered, with AUROC of 0.75 [95% CI 0.73-0.77] and AUPRC of 0.22 [95% CI 0.19-0.25], although our deep learning model (which also included these risk scores) achieved better performance (AUROC 0.92 [95% CI 0.91-0.93] and AUPRC 0.53 [95% CI 0.49-0.57], differences in AUROC $p < 0.05$) (**Figure 5a**). Similar trends were seen when applied to MIMIC-IV data (**Figure 5b**). **Supplement Figure 12** provides a detailed analysis of cases showing disagreement between our deep learning model and the Simplified Acute Physiology Score II predictions.

**Discussion**

In this study, we developed a multimodal deep-learning model incorporating both structured and unstructured data to predict inpatient mortality after the initial 24 hours of an ICU admission. The model demonstrated excellent generalizability based on validation experiments. Intensivists may leverage this technology into their clinical workflow for several use cases, including benchmarking of ICU outcomes, triggering of evaluation for palliative care or withdrawal of care early on, or as an early warning system to better evaluate patients for interventions that may be missing that would be effective in mitigating mortality.

Benchmarking is essential for comparing ICU outcomes fairly, specifically when using standardized mortality ratios, across units, hospitals and over time by adjusting for patient severity and case mix.[39-40] This allows healthcare units with different patient populations to benchmark meaningfully, and thus, not penalizing ICUs (or time periods) that treat sicker patients. Furthermore, it may be used to track performance trends over time and calculate risk-adjusted mortality rather than raw values, identify subgroups where mortality exceed expectations, and audit clinical processes linked to high-risk or high-variance cases. This may provide a more strategic and focused quality improvement review of ICU outcomes. Current validated risk scores, such as Acute Physiology and Chronic Health Evaluation II and National Quality Forum, have been used to calculate standardized mortality ratios;[41] however different prognostic systems can yield different ratios. With ICU outcomes playing an expanding role in reimbursement, referral patterns, and quality improvement initiatives, performance should be evaluated using models with higher accuracy and adjustments for patient heterogeneity. We demonstrated that our model outperformed several risk scores, such as Acute Physiology and Chronic Health Evaluation II and Simplified Acute Physiology Score II, and thus, would potentially be a useful alternative for more accurate benchmarking.

Early warning systems can also predict futility and facilitate discussions on goals of care, thereby sparing patients from uncomfortable interventions.[42-44] These models can identify individuals at high risk of poor outcomes who might otherwise go unrecognized until late in the illness trajectory. Integrating such models into clinical workflows enables proactive—not reactive—consultation with palliative care teams, ensuring that discussions around prognosis, symptom management, and patient values occur before irreversible decline or prolonged nonbeneficial treatment. Early, data-driven identification of patients who may benefit from palliative consultation has been shown to be associated with greater transition to do not resuscitate/intubate and to hospice care as well as decreased healthcare resource utilization.[45]

Our mortality prediction model may also serve as early-warning systems to identify patients who may benefit from timely, potentially life-saving interventions. In resource-limited environments, such as low- and middle-income countries or those affected by public emergencies, such as a pandemic, interventions like mechanical ventilation can be assigned to patients likely to benefit from them.[46] In healthcare systems without apparent shortages, clinicians can preemptively initiate resources, such as dialysis machines or mechanical circulatory support, to patients who would otherwise decompensate in the near future.[47-48] Early sepsis prediction, for example, has been demonstrated using machine learning, which may be used to allocate minimal resources in a timely fashion.[49-50] By calculating risk of inpatient mortality after the initial 24 hour ICU admission, providers may leverage the alerts to enhance situational awareness and provide actionable insight for individualized care.

In addition to generalizability, we addressed other key ethical points regarding AI in our study design, specifically around fairness and bias based on age, sex, race, and ethnicity. Poorly trained AI models may frequently display algorithmic bias, which may lead to predictions that may not be equally accurate across different demographic groups. If providers are unaware of the biases present in these predictive models, then clinical decisions based on recommendations can perpetuate disparities, which may undermine trust and efficacy in diverse patient populations. While it is not always possible to eliminate these biases, researchers should report the presence of it so that users can better utilize model predictions with their own clinical judgement (rather than rely solely on AI-based decisions). In our fairness/bias analyses, we demonstrated a higher false positive rate for older adults. On feature importance analysis, age was consistently one of the top predictors of mortality. We caution clinicians that age alone may not be a sole predictor, but rather act as a surrogate for any unmeasured variables in our datasets, such as frailty. Thus, older patients who are not frail may be more likely to be classified as high risk when, in fact, they are not. Thus, clinicians should look at other factors, such as frailty, to conclude the estimated risk based on their own AI-informed clinical judgement.

Integrating AI models into an institution's electronic health record system is no easy task. Key to its success is a streamlined data harmonization process. A more depth discussion on data harmonization is provided in **Supplement Text 3**. Facilitating data mapping when integrating an AI model built on the MIMIC dataset into a hospital's EHR system requires a structured, standards-based approach to ensure semantic and structural interoperability. This would require a detailed data inventory to identify the model's required input features and their corresponding clinical concepts in the EHR. Leveraging standardized vocabularies such as Logical Observation Identifier Names and Codes, Systematized Nomenclature of Medicine – Clinical Terms, International Classification of Diseases, and RxNorm will aid in consistent representation of labs, diagnoses, and medications.[51] To streamline integration, a converted version of the MIMIC-IV database into Level 7 Fast Healthcare Interoperability Resources (FHIR) standard is available.[52] FHIR serves as a common exchange framework between the MIMIC-based schema and a hospital's EHR, thus developers can use existing FHIR application programming interface to extract and transform clinical data in a scalable, vendor-neutral manner.

Our study has several limitations. First, our model only predicts mortality at one index time point (after the first 24 hours), rather than continuously throughout the ICU stay. A previous study demonstrated that a continuous model outperformed prediction of delirium compared to a 24-hour model.[53] Future steps will be to redesign a more dynamic model for ICU mortality and assess if performance improves compared to our current model. Second, diagnosis of comorbidities was binary and did not include stages of severity. Furthermore, our model was based on retrospective data, which may often be confounded by missing data or unaccounted variables, and, thus, requires further validation in a prospective setting. However, Beth Israel Deaconess Medical Center's dataset is one of the most extensive and contemporary, covering more than 90,000 ICU admissions between 2008 and 2022.[18-19] In addition, MIMIC-IV is one of the first datasets that captured free text notes and radiologic images, which allowed the model to capture valuable information that may not be readily available in other datasets.

Our study demonstrates the utility of a multimodal deep learning model for mortality prediction based on data within the first 24 hours of ICU admission. By including time-series data, static patient information, clinical text notes, and chest X-ray images, we built a mortality prediction model that may improve clinical decision-making and patient outcomes. Future studies may explore the prospective validation of our model.

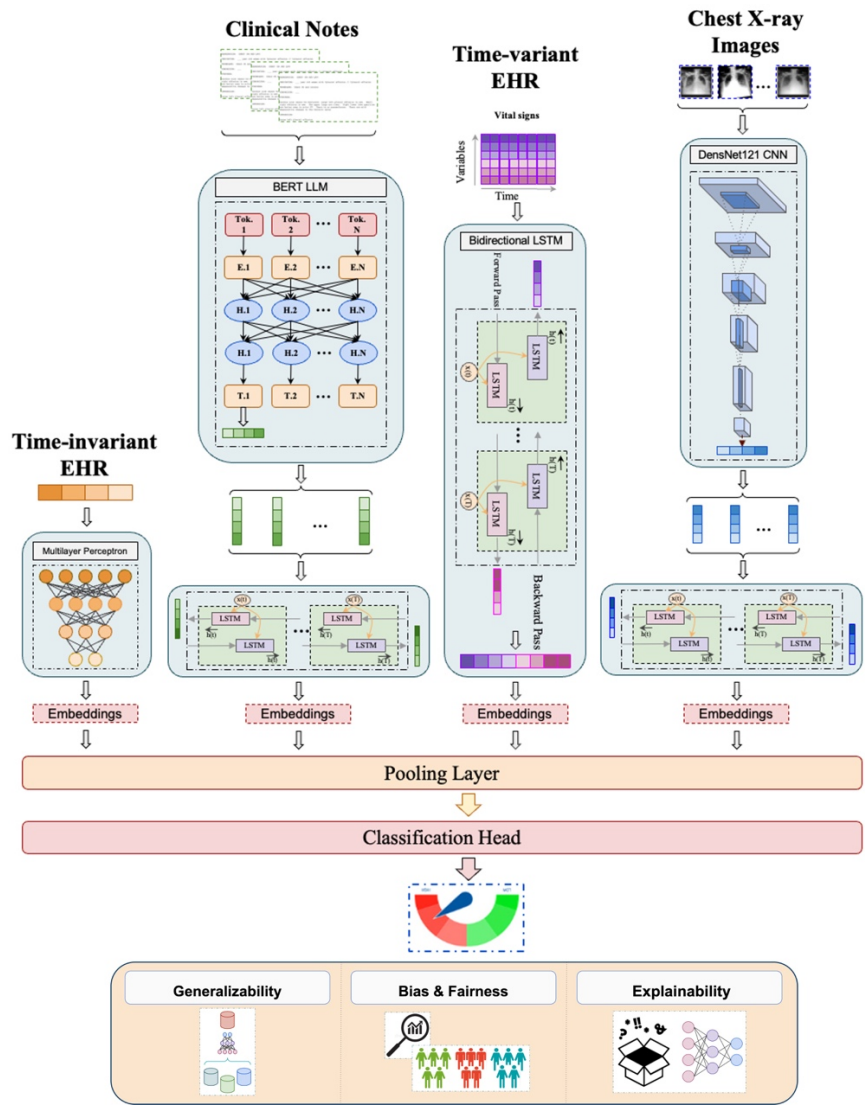

**Figure 1.** Schematic of the multimodal deep learning model for predicting mortality after 24 hours of inpatient data. The architecture of our multimodal deep learning model consists of four component models: 1) a multilayer perceptron for time-invariant variables; 2) a bidirectional long short-term memory model for time-variant data and time-series vital sign variables; 3) a BERT language model for clinical notes; and 4) a DenseNet convolutional neural network for chest X-ray imaging. Model outputs are then unified into a classification model which predicts the outcome. Generalizability, bias/fairness, and explainability were addressed.

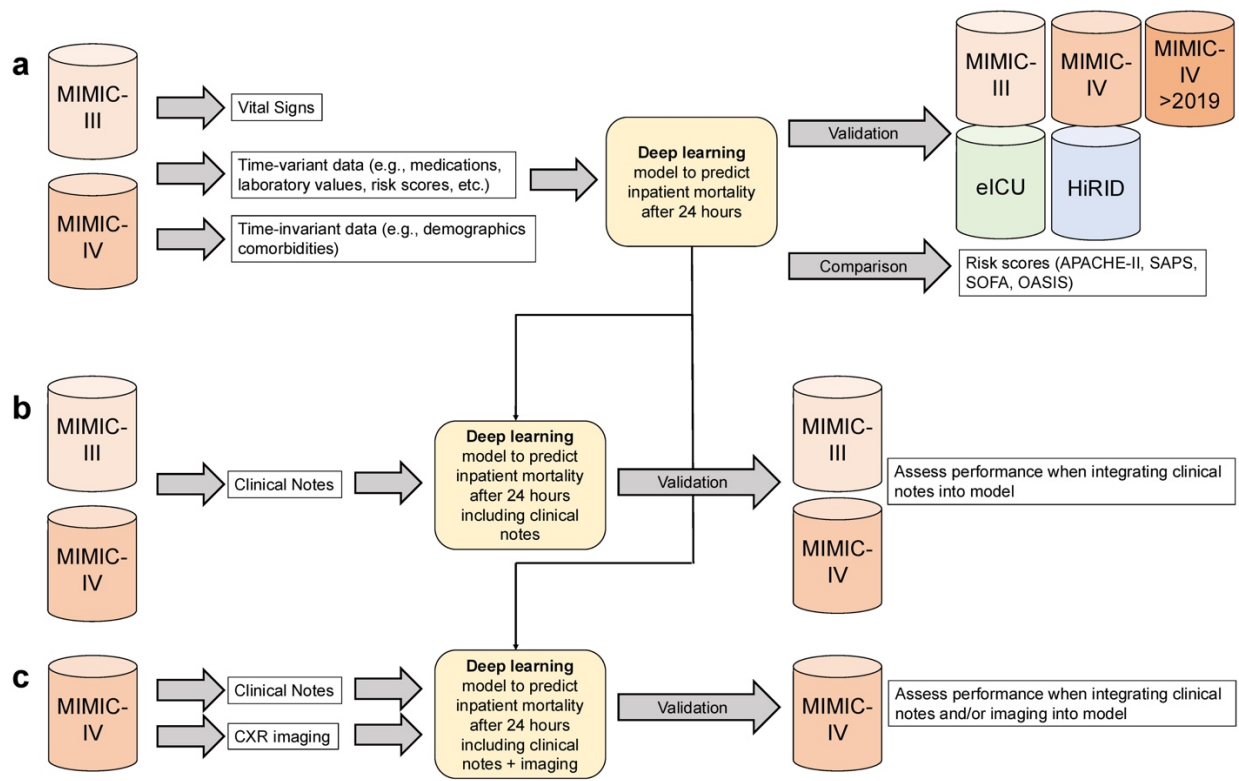

**Figure 2.** Illustration of development and validation approaches: a) using structured time-variant and -invariant data, mortality models were developed on the MIMIC-III and MIMIC-IV datasets and validated internally on MIMIC datasets and externally on eICU and HiRID; b) using structured time-variant, -invariant data, and/or clinical notes, models were developed on the MIMIC-III and MIMIC-IV datasets and validated internally; c) using structured time-variant, -invariant data, clinical notes, and/or chest X-ray images, models were developed on the MIMIC-IV datasets and validated internally

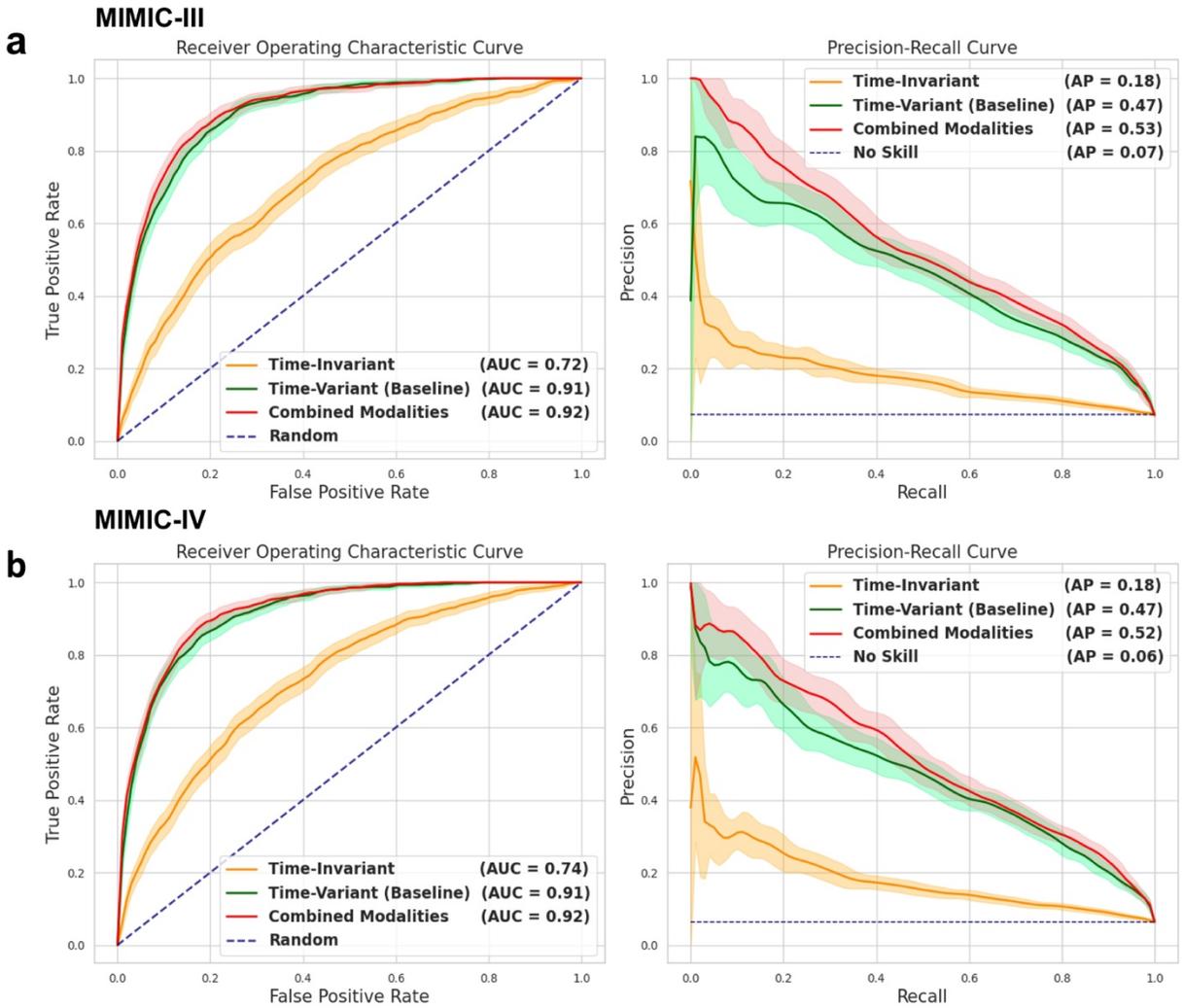

**Figure 3.** Performance of models incorporating time-invariant and -variant features. a) AUROC and AUPRC for the HiRID dataset; b) fairness/bias metrics based on age, race, ethnicity, and sex for HiRID; c) AUROC and AUPRC for the eICU dataset; and d) fairness/bias metrics based on age, race, ethnicity, and sex for the eICU. Abbreviations: AUPRC, area under the precision-recall curve; AUROC, area under the receiver operating characteristics curve.

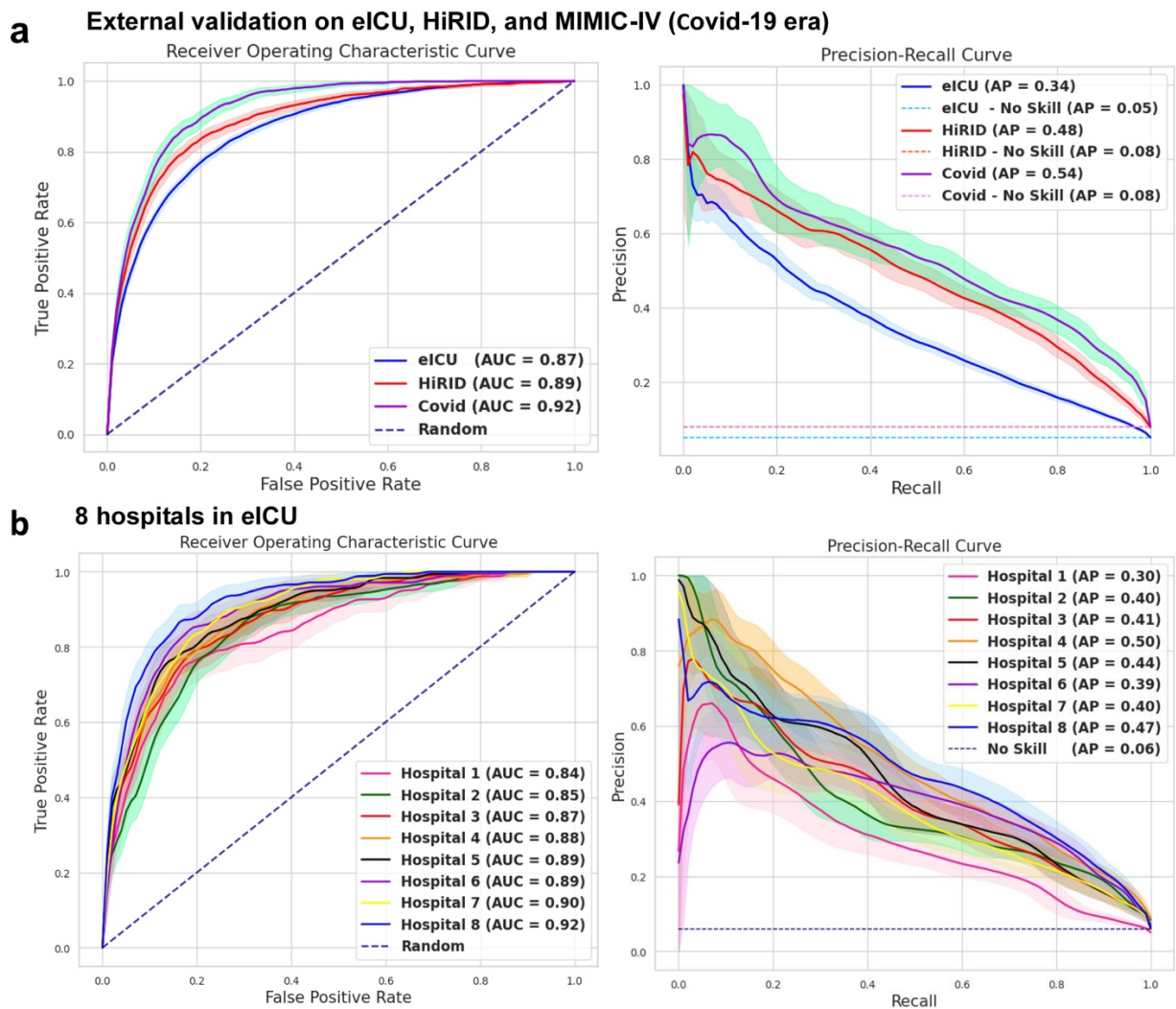

**Figure 4.** External validation the models built on MIMIC datasets using structured data only over HiRID and eICU datasets. a) AUROC and AUPRC of models trained on MIMIC data and tested on the entire HiRID and eICU datasets. b) fairness/bias results assessed for algorithmic bias based on age, sex, gender, race/ethnicity on the test sets. c) AUROC and AUPRC of models trained on the MIMIC data and tested on the 8 separate institutions within the eICU dataset. d) fairness/bias results assessed for algorithmic bias based on age, sex, gender, race/ethnicity on the test sets.

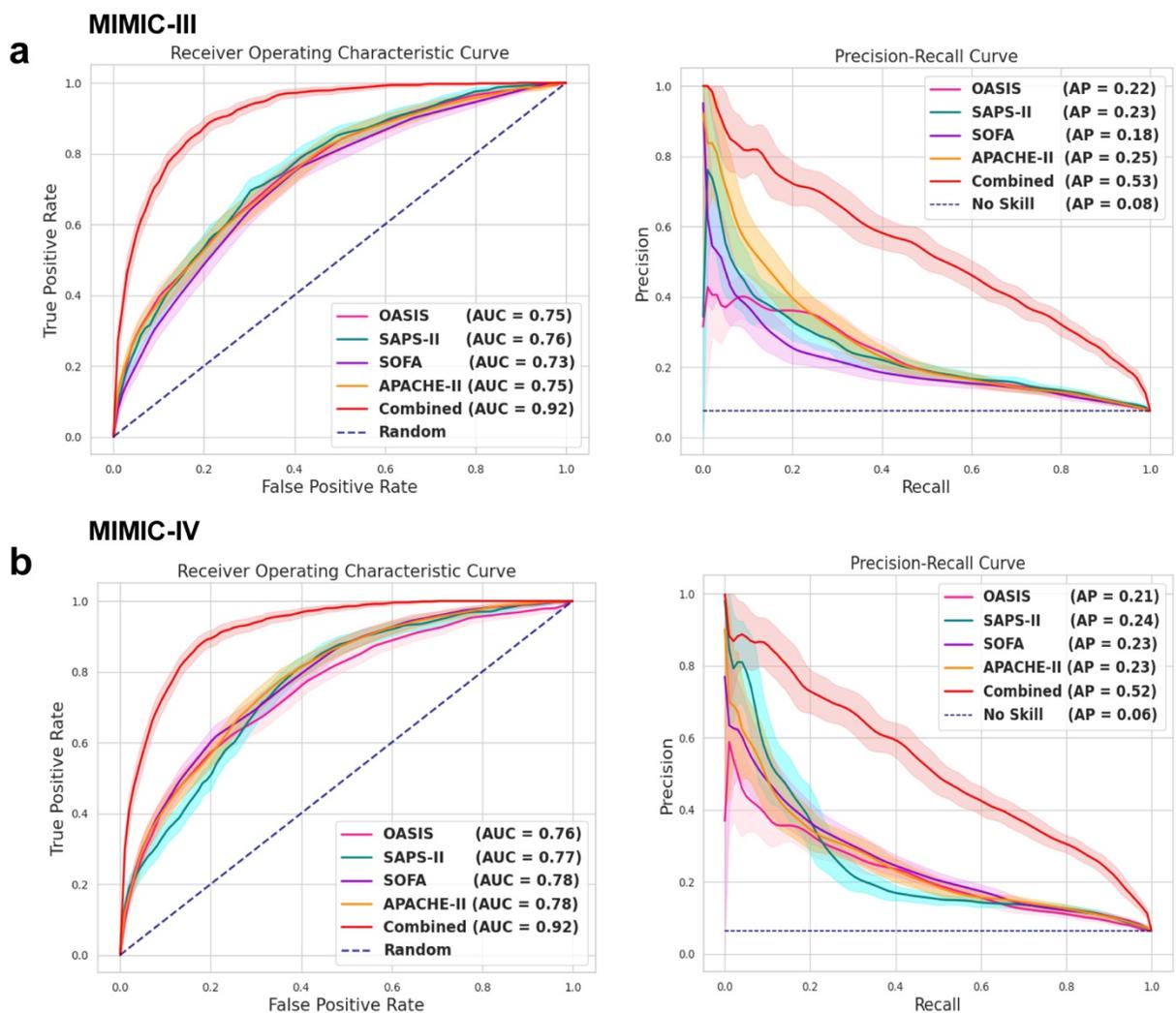

**Figure 5.** a) Area under the receiver operating characteristics curve and area under the precision-recall curve of our multimodal deep learning model combining all data modalities (labeled 'combined', which includes risk score outputs) compared to various risk scores alone, including OASIS, SAPS II, and SOFA using MIMIC-III; and b) using MIMIC-IV.

**Table 1.** Baseline differences between survived and mortality cohorts for each dataset. Differences are calculated by the absolute standardized mean difference.

| | MIMIC-III | | SMD | MIMIC-IV | | SMD | eICU | | SMD | HiRID | | SMD |
|---|---|---|---|---|---|---|---|---|---|---|---|---|
| | Survived | Died | | Survived | Died | | Survived | Died | | Survived | Died | |
| Total, n (%) | 43577 (92.7) | 3406 (7.2) | - | 56299 (93.7) | 3788 (6.3) | - | 125913 (94.8) | 6924 (5.2) | - | 15319 (92.1) | 1323 (7.9) | - |
| Age (years), median [quartiles] | 65.0 [53.0,77.0] | 73.0 [60.0,82.0] | 0.34 | 66.0 [55.0,77.0] | 72.0 [60.0,82.0] | 0.32 | 65.0 [53.0,76.0] | 69.0 [58.0,79.0] | 0.24 | 65.0 [55.0,75.0] | 70.0 [55.0,75.0] | 0.11 |
| Male sex, n (%) | 24738 (56.8) | 1865 (54.8) | 0.04 | 31850 (56.6) | 2063 (54.5) | 0.04 | 68266 (54.2) | 3807 (55.0) | 0.02 | 9774 (63.8) | 823 (62.2) | 0.03 |
| Race, n (%) | | | | | | | | | | | | |
| White | 31405 (72.1) | 2374 (69.7) | 0.05 | 38847 (69.0) | 2355 (62.2) | 0.14 | 96977 (77.0) | 5350 (77.3) | 0.005 | - | - | - |
| Black or African American | 4196 (9.6) | 244 (7.2) | 0.09 | 5948 (10.6) | 352 (9.3) | 0.05 | 14213 (11.3) | 757 (10.9) | 0.01 | - | - | - |
| Asian | 994 (2.3) | 92 (2.7) | 0.02 | 1628 (2.9) | 119 (3.1) | 0.02 | 2091 (1.7) | 132 (1.9) | 0.02 | - | - | - |
| Other/Unknown | 5472 (12.5) | 623 (18.3) | 0.17 | 7704 (13.6) | 857 (22.6) | 0.27 | 7916 (6.3) | 463 (6.7) | 0.01 | - | - | - |
| Hispanic Ethnicity, n (%) | 1510 (3.5) | 73 (2.1) | 0.07 | 2172 (3.9) | 105 (2.8) | 0.06 | 4716 (3.7) | 222 (3.2) | 0.03 | - | - | - |
| Risk Scores, median [quartiles] | | | | | | | | | | | | |
| SAPS-II | 28.0 [19.0,36.0] | 41.0 [30.0,53.0] | 0.92 | 29.0 [21.0,39.0] | 46.0 [33.0,59.0] | 0.98 | 22.0 [15.0,29.0] | 35.0 [24.0,47.0] | 0.91 | 35.0 [24.0,46.0] | 54.0 [43.0,64.0] | 1.12 |
| SOFA | 2.0 [0.0,4.0] | 5.0 [3.0,8.0] | 0.75 | 3.0 [0.0,6.0] | 8.0 [4.0,10.0] | 0.92 | 1.0 [0.0,4.0] | 5.0 [3.0,8.0] | 0.91 | 7.0 [4.0,10.0] | 11.0 [8.0,14.0] | 0.86 |
| OASIS | 19.0 [16.0,21.0] | 22.0 [19.0,29.0] | 0.8 | 19.0 [16.0,21.0] | 23.0 [19.0,29.0] | 0.83 | 15.0 [9.0,19.0] | 19.0 [12.0,23.0] | 0.56 | 18.0 [14.0,22.0] | 26.0 [20.0,30.0] | 0.86 |
| APACHE-II | 5.0 [2.0,9.0] | 11.0 [6.0,15.0] | 0.85 | 10.0 [6.0,14.0] | 16.0 [12.0,21.0] | 0.99 | 8.0 [5.0,11.0] | 14.0 [9.0,19.0] | 0.96 | 12.0 [8.0,16.0] | 19.0 [15.0,23.0] | 1.12 |

Abbreviations: APACHE-II, Acute Physiology and Chronic Health Evaluation II; eICU, Electronic Intensive Care Unit Collaborative Research Database; HiRID, High time-resolution intensive care unit dataset; MIMIC, Medical Information Mart for Intensive Care; OASIS, Oxford Acute Severity of Illness Score; SAPS-II, Simplified Acute Physiology Score II; SMD, absolute standardized mean difference; SOFA, Sequential Organ Failure Assessment

**Table 2.** Performance metrics of each model type based on data used. The first row set corresponds to the model focused on time-invariant and time-variant structured data only. The second row set corresponds to the model incorporating clinical notes. The third row set corresponds to the model integrating both clinical notes and chest x-ray imaging. Sample sizes differ in each model based on data availability. Values are reported with their associated 95% confidence interval in parenthesis.

| | MIMIC-III | | | MIMIC-IV | | |
| --- | --- | --- | --- | --- | --- | --- |
| | **AUROC** | **AUPRC** | **Brier score** | **AUROC** | **AUPRC** | **Brier score** |
| **Model** (only structured data) | | | | | | |
| Time-invariant data only | 0.717 (0.697-0.737) | 0.178 (0.156-0.204) | 0.145 (0.143-0.147) | 0.736 (0.719-0.755) | 0.181 (0.158-0.206) | 0.389 (0.385-0.393) |
| Time-variant data only | 0.906 (0.895-0.915) | 0.471 (0.429-0.513) | 0.261 (0.255-0.267) | 0.912 (0.903-0.921) | 0.473 (0.434-0.512) | 0.282 (0.277-0.289) |
| Combined modalities | 0.916 (0.904-0.925) | 0.531 (0.491-0.568) | 0.191 (0.185-0.197) | 0.921 (0.912-0.930) | 0.518 (0.476-0.558) | 0.064 (0.061-0.067) |
| **Model** (+clinical notes) | | | | | | |
| Clinical notes only | 0.859 (0.843-0.873) | 0.391 (0.350-0.430) | 0.136 (0.130-0.141) | 0.819 (0.803-0.834) | 0.282 (0.249-0.315) | 0.215 (0.209-0.221) |
| Time-variant data only | 0.905 (0.894-0.914) | 0.468 (0.426-0.511) | 0.267 (0.261-0.273) | 0.905 (0.895-0.915) | 0.482 (0.443-0.520) | 0.307 (0.300-0.314) |
| Combined modalities | 0.913 (0.903-0.923) | 0.512 (0.471-0.553) | 0.080 (0.077-0.084) | 0.913 (0.903-0.923) | 0.507 (0.467-0.547) | 0.058 (0.055-0.061) |
| **Model** (+clinical notes and imaging) | *no imaging available in MIMIC-III dataset | | | | | |
| Chest X-ray imaging only | - | - | - | 0.758 (0.721-0.794) | 0.252 (0.196-0.313) | 0.253 (0.244-0.264) |
| Clinical notes only | - | - | - | 0.758 (0.721-0.794) | 0.288 (0.231-0.354) | 0.271 (0.258-0.285) |
| Time-invariant data only | - | - | - | 0.718 (0.681-0.755) | 0.222 (0.172-0.282) | 0.404 (0.394-0.414) |
| Time-variant data only | - | - | - | 0.874 (0.847-0.897) | 0.434 (0.358-0.506) | 0.374 (0.359-0.390) |
| Combined modalities | - | - | - | 0.891 (0.866-0.913) | 0.482 (0.398-0.555) | 0.173 (0.161-0.187) |

Abbreviations: AUPRC, area under the precision-recall curve; AUROC, area under the receiver operating characteristics curve; MIMIC, Medical Information Mart for Intensive Care

# Supplementary Materials

Supplementary to: **Development and external validation of a multimodal artificial intelligence mortality prediction model of critically ill patients using multicenter data**

Behrooz Mamandipoor, BS[1,2]; Chun-Nan Hsu, PhD[2]; Martin Krause, MD[2,3]; Ulrich H. Schmidt, MD, PhD, MBA[3]; Rodney A. Gabriel, MD, MAS[1,2]

[1]Department of Biomedical Informatics, University of California, San Diego Health, La Jolla, CA, USA
[2]Division of Perioperative Informatics, Department of Anesthesiology, University of California, San Diego, La Jolla, CA, USA
[3]Division of Critical Care Medicine, Department of Anesthesiology, University of California, San Diego, La Jolla, CA, USA

**Corresponding author:**
Rodney A. Gabriel, MD, MAS
(858) 663 7747
ragabriel@health.ucsd.edu

**Funding disclosures:** Departmental internal funding through the Division of Perioperative Informatics

**Conflicts of interest:** none

# TABLE OF CONTENTS

**Supplement Figure 1:** *Patient Inclusion and Exclusion Criteria* for a) eICU; b) HiRID; c) MIMIC-III; and d) MIMIC-IV.

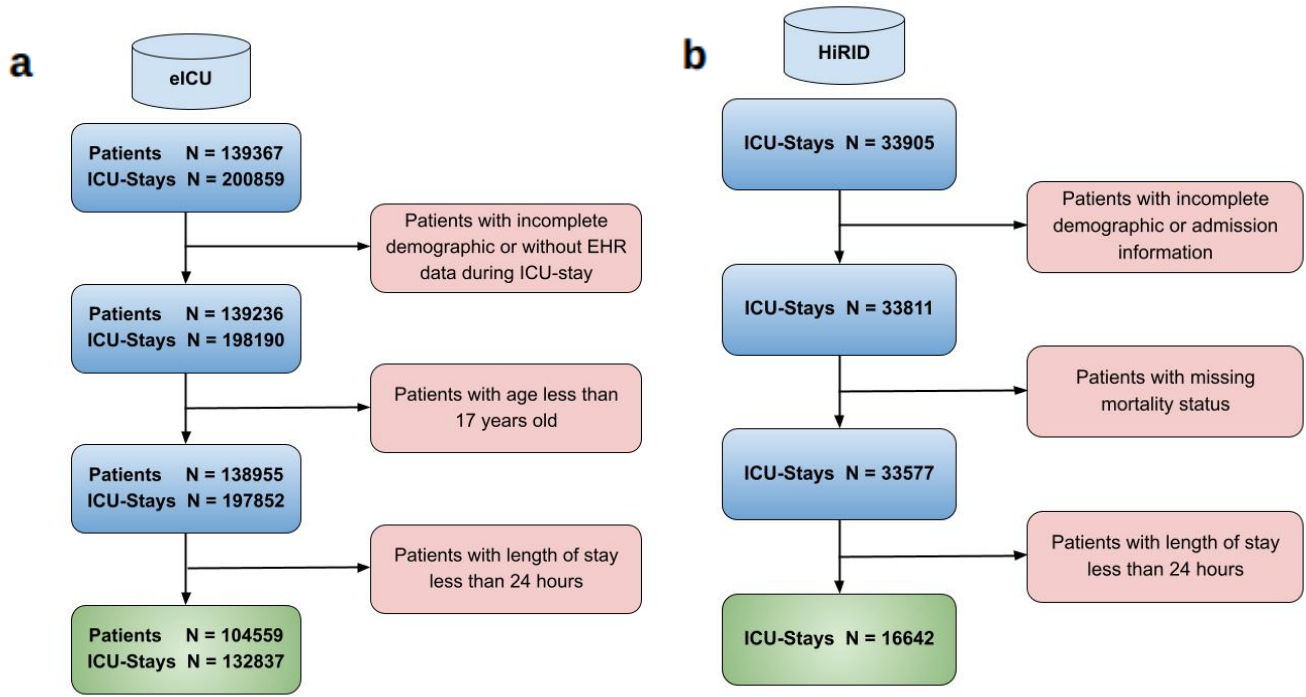

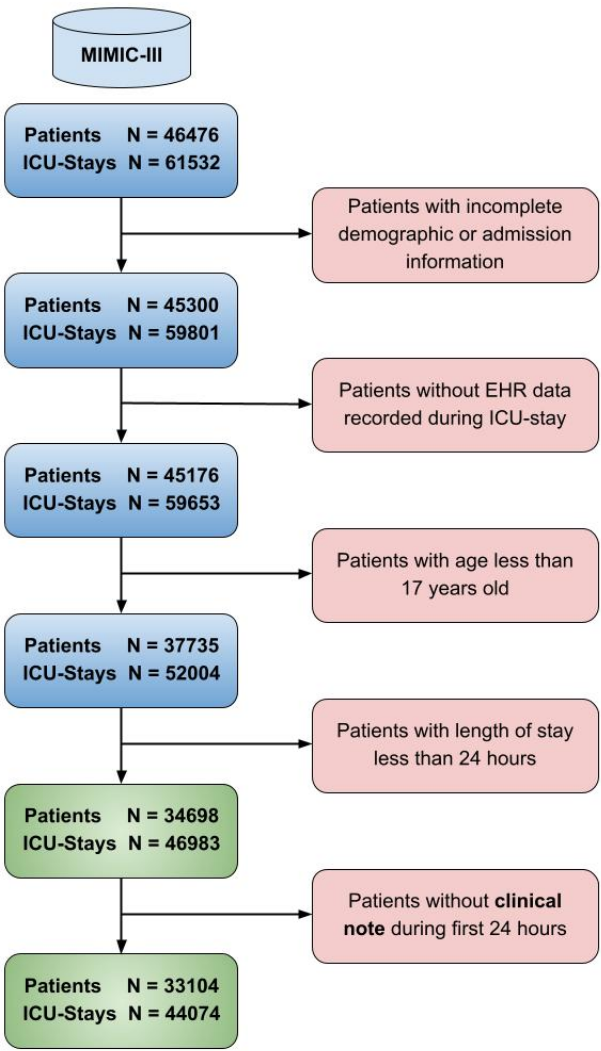

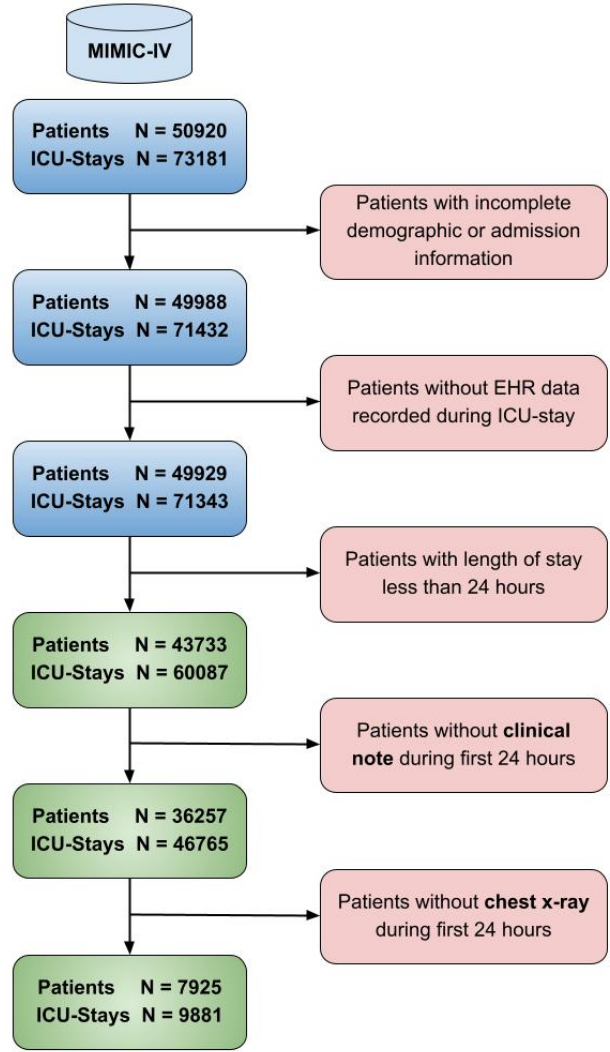

c

d

**Supplement Table 1:** *Distribution of Data Points for eICU*

| | Overall | Survived | Dead | P-Value |
|---|---|---|---|---|
| age, median [Q1,Q3] | 66.0 [54.0,76.0] | 65.0 [53.0,76.0] | 69.0 [58.0,79.0] | <0.001 |
| gender, n (%) Unknown | 56 (0.0) | 53 (0.0) | 3 (0.0) | 0.459 |
| gender, n (%) Male | 72073 (54.3) | 68266 (54.2) | 3807 (55.0) | 0.459 |
| gender, n (%) Female | 60708 (45.7) | 57594 (45.7) | 3114 (45.0) | 0.459 |
| ethnicity, n (%) African American | 14970 (11.3) | 14213 (11.3) | 757 (10.9) | 0.074 |
| ethnicity, n (%) Asian | 2223 (1.7) | 2091 (1.7) | 132 (1.9) | 0.074 |
| ethnicity, n (%) Caucasian | 102327 (77.0) | 96977 (77.0) | 5350 (77.3) | 0.074 |
| ethnicity, n (%) Hispanic | 4938 (3.7) | 4716 (3.7) | 222 (3.2) | 0.074 |
| ethnicity, n (%) Native American | 873 (0.7) | 823 (0.7) | 50 (0.7) | 0.074 |
| ethnicity, n (%) Other/Unknown | 7506 (5.7) | 7093 (5.6) | 413 (6.0) | 0.074 |
| unitdischargeoffset, median [Q1,Q3] | 55.6 [38.2,97.3] | 54.4 [37.9,95.2] | 85.0 [46.1,169.4] | <0.001 |
| ST1 (ECG ST elevation), median [Q1,Q3] | 0.0 [-0.1,0.2] | 0.0 [-0.1,0.2] | 0.0 [-0.2,0.1] | <0.001 |
| ST2 (ECG ST elevation), median [Q1,Q3] | 0.0 [-0.3,0.3] | 0.0 [-0.3,0.3] | 0.0 [-0.4,0.2] | <0.001 |
| ST3 (ECG ST elevation), median [Q1,Q3] | 0.0 [-0.2,0.2] | 0.0 [-0.2,0.2] | 0.0 [-0.3,0.2] | <0.001 |
| Heart Rate, median [Q1,Q3] | 83.8 [72.0,97.2] | 83.4 [71.8,96.6] | 92.0 [77.3,107.2] | <0.001 |
| SpO2, median [Q1,Q3] | 97.3 [95.3,99.1] | 97.3 [95.3,99.0] | 97.4 [94.6,99.6] | 0.017 |
| Oxygen Saturation, median [Q1,Q3] | 97.0 [95.0,99.0] | 97.0 [95.0,99.0] | 97.5 [95.0,99.7] | <0.001 |
| Respiratory Rate, median [Q1,Q3] | 18.8 [16.0,22.5] | 18.7 [15.9,22.3] | 20.9 [17.0,25.8] | <0.001 |
| Temperature (C), median [Q1,Q3] | 36.9 [36.5,37.3] | 36.9 [36.6,37.3] | 36.7 [35.9,37.3] | <0.001 |
| Non-Invasive BP Mean, median [Q1,Q3] | 79.2 [69.5,91.0] | 79.8 [70.0,91.3] | 74.0 [65.0,85.0] | <0.001 |
| Non-Invasive BP Systolic, median [Q1,Q3] | 118.0 [104.0,135.0] | 118.5 [104.5,135.2] | 109.5 [97.0,125.8] | <0.001 |
| Non-Invasive BP Diastolic, median [Q1,Q3] | 64.0 [55.2,74.0] | 64.0 [55.5,74.3] | 60.0 [52.0,70.0] | <0.001 |
| Glucose, median [Q1,Q3] | 135.0 [109.0,175.0] | 135.0 [109.0,173.0] | 143.0 [110.0,194.0] | <0.001 |
| Creatinine, median [Q1,Q3] | 1.1 [0.8,1.8] | 1.0 [0.8,1.7] | 1.6 [1.0,2.6] | <0.001 |
| Base Excess, median [Q1,Q3] | -1.4 [-5.0,2.0] | -1.0 [-4.7,2.3] | -4.7 [-9.9,0.6] | <0.001 |
| BUN, median [Q1,Q3] | 20.0 [13.0,35.0] | 20.0 [13.0,34.0] | 30.0 [19.0,48.0] | <0.001 |
| Anion Gap, median [Q1,Q3] | 11.0 [8.0,14.0] | 11.0 [8.0,14.0] | 13.0 [10.0,17.0] | <0.001 |
| Bicarbonate, median [Q1,Q3] | 23.0 [20.0,26.0] | 23.0 [20.1,26.0] | 20.9 [17.0,24.5] | <0.001 |
| Lactate, median [Q1,Q3] | 1.9 [1.2,3.3] | 1.8 [1.1,2.9] | 3.5 [1.9,6.9] | <0.001 |
| Hemoglobin, median [Q1,Q3] | 10.4 [8.8,12.2] | 10.4 [8.8,12.2] | 10.1 [8.6,12.0] | <0.001 |
| Hematocrit, median [Q1,Q3] | 31.5 [26.9,36.6] | 31.6 [27.0,36.6] | 30.9 [26.3,36.4] | <0.001 |
| pH, median [Q1,Q3] | 7.4 [7.3,7.4] | 7.4 [7.3,7.4] | 7.3 [7.2,7.4] | <0.001 |
| Bilirubin, Direct, median [Q1,Q3] | 0.2 [0.1,0.6] | 0.2 [0.1,0.6] | 0.4 [0.2,1.1] | <0.001 |
| pO2, median [Q1,Q3] | 101.0 [77.0,149.0] | 101.4 [77.0,148.0] | 98.0 [72.0,155.0] | <0.001 |
| pCO2, median [Q1,Q3] | 40.0 [34.4,47.5] | 40.4 [35.0,47.5] | 39.0 [32.0,47.4] | <0.001 |
| AST, median [Q1,Q3] | 33.0 [20.0,72.0] | 32.0 [19.0,66.0] | 67.0 [34.0,166.0] | <0.001 |
| ALT, median [Q1,Q3] | 27.0 [17.0,53.0] | 27.0 [16.0,50.0] | 42.0 [22.0,112.0] | <0.001 |
| WBC x 1000, median [Q1,Q3] | 11.0 [7.9,15.2] | 10.8 [7.9,14.9] | 13.3 [8.7,19.0] | <0.001 |
| RBC, median [Q1,Q3] | 3.6 [3.1,4.1] | 3.6 [3.1,4.2] | 3.4 [2.9,4.1] | <0.001 |
| Potassium, median [Q1,Q3] | 4.0 [3.7,4.5] | 4.0 [3.7,4.5] | 4.1 [3.6,4.7] | <0.001 |
| Sodium, median [Q1,Q3] | 139.0 [135.0,142.0] | 139.0 [135.0,141.0] | 139.0 [135.0,143.0] | <0.001 |
| Chloride, median [Q1,Q3] | 105.0 [101.0,109.0] | 105.0 [101.0,109.0] | 105.0 [100.0,110.0] | <0.001 |
| Magnesium, median [Q1,Q3] | 1.9 [1.7,2.2] | 1.9 [1.7,2.2] | 2.0 [1.7,2.3] | <0.001 |
| Phosphate, median [Q1,Q3] | 3.3 [2.5,4.2] | 3.2 [2.5,4.1] | 3.9 [2.8,5.3] | <0.001 |
| FiO2, median [Q1,Q3] | 45.0 [36.0,65.0] | 40.0 [35.0,60.0] | 60.0 [40.0,100.0] | <0.001 |
| PEEP, median [Q1,Q3] | 5.0 [5.0,6.0] | 5.0 [5.0,5.0] | 5.0 [5.0,8.0] | <0.001 |
| Tidal Volume, median [Q1,Q3] | 500.0 [426.0,550.0] | 500.0 [429.0,550.0] | 500.0 [420.0,539.5] | <0.001 |
| UrineOutput_IO, median [Q1,Q3] | 100.0 [45.0,250.0] | 102.0 [50.0,258.0] | 50.0 [20.0,137.0] | <0.001 |
| PaO2/FiO2, median [Q1,Q3] | 2.5 [1.7,3.6] | 2.5 [1.7,3.6] | 2.0 [1.2,3.2] | <0.001 |
| SIRS, median [Q1,Q3] | 1.0 [0.0,2.0] | 1.0 [0.0,2.0] | 2.0 [1.0,2.0] | <0.001 |
| Shock_Index, median [Q1,Q3] | 0.7 [0.6,0.9] | 0.7 [0.6,0.8] | 0.8 [0.7,1.1] | <0.001 |
| SOFA, median [Q1,Q3] | 1.0 [0.0,4.0] | 1.0 [0.0,4.0] | 5.0 [3.0,8.0] | <0.001 |
| SAPSII, median [Q1,Q3] | 23.0 [15.0,31.0] | 22.0 [15.0,29.0] | 35.0 [24.0,47.0] | <0.001 |
| OASIS, median [Q1,Q3] | 15.0 [9.0,19.0] | 15.0 [9.0,19.0] | 19.0 [12.0,23.0] | <0.001 |
| Total GCS, median [Q1,Q3] | 15.0 [11.0,15.0] | 15.0 [12.0,15.0] | 9.0 [4.0,14.0] | <0.001 |

**Supplement Table 2:** *Distribution of Data Points for HiRID*

| | Overall | Survived | Dead | P-Value |
|---|---|---|---|---|
| age, median [Q1,Q3] | 65.0 [55.0,75.0] | 65.0 [55.0,75.0] | 70.0 [55.0,75.0] | <0.001 |
| gender, n (%) Male | 10597 (63.7) | 9774 (63.8) | 823 (62.2) | 0.259 |
| gender, n (%) Female | 6045 (36.3) | 5545 (36.2) | 500 (37.8) | 0.259 |
| icuLos_h, median [Q1,Q3] | 48.0 [31.2,98.4] | 48.0 [31.2,96.0] | 62.4 [38.4,122.4] | <0.001 |
| ST1 (ECG ST elevation), median [Q1,Q3] | 0.0 [-0.2,0.3] | 0.0 [-0.2,0.3] | 0.0 [-0.3,0.2] | <0.001 |
| ST2 (ECG ST elevation), median [Q1,Q3] | 0.0 [-0.1,0.3] | 0.1 [-0.1,0.3] | 0.0 [-0.2,0.3] | <0.001 |
| ST3 (ECG ST elevation), median [Q1,Q3] | 0.0 [-0.1,0.2] | 0.0 [-0.1,0.2] | 0.0 [-0.1,0.2] | 0.213 |
| Heart Rate, median [Q1,Q3] | 84.0 [71.0,96.0] | 84.0 [71.0,96.0] | 87.0 [71.0,102.0] | <0.001 |
| SpO2, median [Q1,Q3] | 97.0 [95.0,99.0] | 97.0 [95.0,99.0] | 98.0 [95.0,100.0] | 0.57 |
| Oxygen Saturation_SO2, median [Q1,Q3] | 98.0 [96.0,99.0] | 98.0 [96.0,99.0] | 97.7 [95.1,99.0] | <0.001 |
| Respiratory Rate, median [Q1,Q3] | 17.0 [13.0,22.0] | 17.0 [13.0,22.0] | 16.0 [12.0,22.1] | <0.001 |
| Temperature Central, median [Q1,Q3] | 37.1 [36.5,37.6] | 37.1 [36.6,37.6] | 37.0 [36.0,37.6] | <0.001 |
| Invasive systolic arterial pressure, median [Q1,Q3] | 113.0 [95.0,133.0] | 113.0 [96.0,133.0] | 104.0 [87.0,126.0] | <0.001 |
| Invasive diastolic arterial pressure, median [Q1,Q3] | 55.0 [48.0,64.0] | 55.0 [48.0,64.0] | 53.0 [46.0,62.0] | <0.001 |
| Invasive mean arterial pressure, median [Q1,Q3] | 74.0 [64.0,86.0] | 75.0 [65.0,87.0] | 70.0 [60.0,83.0] | <0.001 |
| Glucose, median [Q1,Q3] | 7.9 [6.4,9.8] | 7.9 [6.4,9.8] | 8.1 [6.5,10.1] | <0.001 |
| Creatinine, median [Q1,Q3] | 394.0 [153.0,1104.2] | 383.0 [150.0,1038.0] | 547.0 [176.0,1627.0] | <0.001 |
| Base Excess, median [Q1,Q3] | -1.2 [-4.0,1.1] | -1.0 [-3.6,1.2] | -3.5 [-7.6,-0.3] | <0.001 |
| BUN, median [Q1,Q3] | 3.4 [2.2,5.4] | 3.3 [2.2,5.3] | 4.2 [2.6,6.8] | <0.001 |
| Bicarbonate, median [Q1,Q3] | 23.1 [20.7,25.2] | 23.3 [21.0,25.3] | 21.1 [17.8,23.9] | <0.001 |
| Lactate, median [Q1,Q3] | 1.6 [1.0,2.7] | 1.5 [1.0,2.5] | 2.7 [1.5,5.2] | <0.001 |
| Hemoglobin, median [Q1,Q3] | 102.0 [90.0,118.0] | 102.0 [90.0,118.0] | 106.0 [92.0,125.0] | <0.001 |
| Carboxy Hemoglobin, median [Q1,Q3] | 1.4 [1.0,1.7] | 1.4 [1.1,1.7] | 1.2 [0.9,1.6] | <0.001 |
| pH, median [Q1,Q3] | 7.4 [7.4,7.5] | 7.4 [7.4,7.5] | 7.4 [7.3,7.4] | <0.001 |
| Bilirubin, Direct, median [Q1,Q3] | 8.8 [5.4,18.5] | 8.8 [5.4,17.9] | 9.5 [5.4,23.4] | 0.001 |
| PO2, median [Q1,Q3] | 98.4 [79.5,129.0] | 98.6 [80.0,128.0] | 98.0 [77.6,131.0] | 0.058 |
| pCO2, median [Q1,Q3] | 35.3 [31.5,39.8] | 35.4 [31.7,39.9] | 34.6 [30.0,39.5] | <0.001 |
| AST, median [Q1,Q3] | 57.0 [28.0,186.0] | 54.0 [27.0,163.0] | 100.0 [38.0,365.0] | <0.001 |
| ALT, median [Q1,Q3] | 35.0 [18.0,101.0] | 34.0 [18.0,92.0] | 54.0 [22.5,177.0] | <0.001 |
| Calcium, median [Q1,Q3] | 2.0 [1.9,2.1] | 2.0 [1.9,2.1] | 2.1 [1.9,2.2] | 0.028 |
| Potassium, median [Q1,Q3] | 4.1 [3.7,4.5] | 4.1 [3.7,4.5] | 4.1 [3.7,4.6] | 0.025 |
| Sodium, median [Q1,Q3] | 137.0 [134.0,140.0] | 137.0 [134.0,140.0] | 138.0 [135.0,141.0] | <0.001 |
| Chloride, median [Q1,Q3] | 108.0 [105.0,111.0] | 108.0 [105.0,111.0] | 109.0 [105.0,112.0] | <0.001 |
| Magnesium, median [Q1,Q3] | 0.8 [0.7,0.9] | 0.8 [0.7,0.9] | 0.8 [0.7,1.0] | <0.001 |
| Phosphate, median [Q1,Q3] | 1.1 [0.9,1.4] | 1.1 [0.9,1.4] | 1.2 [0.9,1.6] | <0.001 |
| FIO2, median [Q1,Q3] | 40.6 [39.8,50.4] | 40.4 [39.8,50.2] | 45.1 [39.9,59.9] | <0.001 |
| PEEP, median [Q1,Q3] | 5.1 [4.8,7.5] | 5.1 [4.8,7.3] | 5.2 [4.9,8.0] | <0.001 |
| Tidal Volume, median [Q1,Q3] | 530.0 [464.0,608.0] | 530.0 [463.0,609.0] | 528.0 [465.0,602.0] | 0.003 |
| UrineOutput_IO, median [Q1,Q3] | 74.3 [43.5,153.8] | 75.0 [45.0,155.2] | 62.3 [31.0,135.0] | <0.001 |
| PaO2/FiO2, median [Q1,Q3] | 2.2 [1.4,3.2] | 2.1 [1.4,3.2] | 2.3 [1.5,3.3] | <0.001 |
| SIRS, median [Q1,Q3] | 1.0 [0.0,2.0] | 1.0 [0.0,2.0] | 1.0 [1.0,2.0] | <0.001 |
| Shock_Index, median [Q1,Q3] | 0.7 [0.6,0.9] | 0.7 [0.6,0.9] | 0.8 [0.6,1.1] | <0.001 |
| SOFA, median [Q1,Q3] | 8.0 [4.0,11.0] | 7.0 [4.0,10.0] | 11.0 [8.0,14.0] | <0.001 |
| SAPSII, median [Q1,Q3] | 37.0 [25.0,48.0] | 35.0 [24.0,46.0] | 54.0 [43.0,64.0] | <0.001 |
| OASIS, median [Q1,Q3] | 19.0 [15.0,23.0] | 18.0 [14.0,22.0] | 26.0 [20.0,30.0] | <0.001 |
| Total GCS, median [Q1,Q3] | 14.0 [8.0,15.0] | 14.0 [10.0,15.0] | 6.0 [3.0,12.0] | <0.001 |

**Supplement Table 3:** *Distribution of Data Points for MIMIC-III*

| | Overall | Survived | Dead | P-Value |
|---|---|---|---|---|
| ICU-Stays | 46983 | 43577 | 3406 | |
| AGE, median [Q1,Q3] | 66.0 [53.0,78.0] | 65.0 [53.0,77.0] | 73.0 [60.0,82.0] | <0.001 |
| GENDER, n (%) - Male | 26603 (56.6) | 24738 (56.8) | 1865 (54.8) | 0.024 |
| GENDER, n (%) - Female | 20380 (43.4) | 18839 (43.2) | 1541 (45.2) | |
| RACE, n (%) - Asian | 1086 (2.3) | 994 (2.3) | 92 (2.7) | <0.001 |
| RACE, n (%) - Black/African American | 4440 (9.5) | 4196 (9.6) | 244 (7.2) | |
| RACE, n (%) - Hispanic/Latino | 1583 (3.4) | 1510 (3.5) | 73 (2.1) | |
| RACE, n (%) - Native American | 23 (0.0) | 20 (0.0) | 3 (0.1) | |
| RACE, n (%) - Other | 1190 (2.5) | 1109 (2.5) | 81 (2.4) | |
| RACE, n (%) - Unknown | 4882 (10.4) | 4343 (10.0) | 539 (15.8) | |
| RACE, n (%) - White | 33779 (71.9) | 31405 (72.1) | 2374 (69.7) | |
| ETHNICITY, n (%) - Hispanic | 1583 (3.4) | 1510 (3.5) | 73 (2.1) | <0.001 |
| ETHNICITY, n (%) - Non Hispanic | 45400 (96.6) | 42067 (96.5) | 3333 (97.9) | |
| ICU_LOS_H, median [Q1,Q3] | 56.7 [34.6,109.8] | 54.5 [33.7,100.8] | 114.0 [52.3,235.2] | <0.001 |
| Number of Notes, median [Q1,Q3] | 4.0 [3.0,6.0] | 4.0 [3.0,6.0] | 6.0 [4.0,8.0] | <0.001 |
| Heart Rate, median [Q1,Q3] | 84.0 [73.0,97.0] | 84.0 [72.7,96.0] | 90.0 [76.0,105.0] | <0.001 |
| SpO2, median [Q1,Q3] | 98.0 [96.0,99.8] | 98.0 [96.0,99.7] | 98.0 [95.0,100.0] | <0.001 |
| Oxygen Saturation, median [Q1,Q3] | 96.0 [84.5,98.0] | 97.0 [85.0,98.0] | 95.0 [82.0,98.0] | <0.001 |
| Respiratory Rate, median [Q1,Q3] | 18.0 [15.0,22.0] | 18.0 [15.0,22.0] | 20.0 [16.0,25.0] | <0.001 |
| Temperature, median [Q1,Q3] | 36.9 [36.4,37.5] | 36.9 [36.4,37.5] | 36.8 [36.1,37.5] | <0.001 |
| Non Invasive Blood Pressure mean, median [Q1,Q3] | 75.0 [66.0,86.0] | 75.3 [66.0,86.0] | 70.0 [61.9,81.0] | <0.001 |
| Non Invasive Blood Pressure systolic, median [Q1,Q3] | 116.0 [103.0,133.0] | 117.0 [103.8,133.0] | 109.0 [96.0,126.0] | <0.001 |
| Non Invasive Blood Pressure diastolic, median [Q1,Q3] | 60.0 [50.0,70.0] | 60.0 [51.0,70.0] | 55.0 [46.0,66.0] | <0.001 |
| Glucose, median [Q1,Q3] | 129.0 [106.0,162.0] | 128.5 [106.0,160.5] | 140.0 [109.0,185.0] | <0.001 |
| Creatinine, median [Q1,Q3] | 1.0 [0.7,1.6] | 1.0 [0.7,1.5] | 1.4 [0.9,2.4] | <0.001 |
| Base Excess, median [Q1,Q3] | -1.0 [-4.0,1.0] | 0.0 [-3.0,1.0] | -3.0 [-8.0,0.0] | <0.001 |
| BUN, median [Q1,Q3] | 20.0 [13.0,34.0] | 19.0 [13.0,32.0] | 32.0 [19.0,52.0] | <0.001 |
| Anion Gap, median [Q1,Q3] | 14.0 [11.0,16.0] | 13.0 [11.0,16.0] | 16.0 [13.0,19.0] | <0.001 |
| Bicarbonate, median [Q1,Q3] | 24.0 [21.0,26.0] | 24.0 [21.0,26.0] | 21.0 [17.0,25.0] | <0.001 |
| Lactate, median [Q1,Q3] | 2.0 [1.3,3.2] | 1.9 [1.3,2.9] | 3.0 [1.8,5.4] | <0.001 |
| Hemoglobin, median [Q1,Q3] | 10.2 [9.1,11.5] | 10.2 [9.1,11.5] | 10.2 [9.0,11.5] | 0.001 |
| Hematocrit, median [Q1,Q3] | 30.3 [27.1,34.0] | 30.3 [27.1,34.0] | 30.4 [27.0,34.5] | 0.001 |
| pH, median [Q1,Q3] | 7.4 [7.3,7.4] | 7.4 [7.3,7.4] | 7.3 [7.2,7.4] | <0.001 |
| Bilirubin, Direct, median [Q1,Q3] | 0.4 [0.1,1.3] | 0.4 [0.1,1.2] | 0.7 [0.2,2.6] | <0.001 |
| pO2, median [Q1,Q3] | 129.0 [92.0,195.0] | 131.0 [93.0,199.0] | 111.0 [80.0,165.0] | <0.001 |
| pCO2, median [Q1,Q3] | 40.0 [36.0,46.0] | 40.0 [36.0,46.0] | 38.5 [33.0,46.0] | <0.001 |
| AST, median [Q1,Q3] | 47.0 [25.0,111.0] | 44.0 [25.0,104.0] | 70.0 [33.0,169.0] | <0.001 |
| ALT, median [Q1,Q3] | 34.0 [18.0,82.0] | 33.0 [18.0,78.0] | 43.0 [22.0,108.0] | <0.001 |
| WBC, median [Q1,Q3] | 10.5 [7.1,14.6] | 10.4 [7.1,14.4] | 11.6 [6.8,17.5] | <0.001 |
| RBC, median [Q1,Q3] | 3.5 [3.0,4.0] | 3.5 [3.0,4.0] | 3.5 [3.0,4.1] | 0.493 |
| Potassium, median [Q1,Q3] | 4.1 [3.8,4.6] | 4.1 [3.8,4.6] | 4.2 [3.7,4.7] | <0.001 |
| Sodium, median [Q1,Q3] | 138.0 [136.0,141.0] | 138.0 [136.0,141.0] | 139.0 [135.0,142.0] | <0.001 |
| Chloride, median [Q1,Q3] | 106.0 [102.0,110.0] | 106.0 [102.0,110.0] | 106.0 [100.0,110.0] | <0.001 |
| Magnesium, median [Q1,Q3] | 2.0 [1.8,2.2] | 2.0 [1.8,2.2] | 2.0 [1.8,2.3] | <0.001 |
| Phosphate, median [Q1,Q3] | 3.4 [2.8,4.3] | 3.4 [2.7,4.2] | 3.9 [3.0,5.1] | <0.001 |
| FiO2, median [Q1,Q3] | 50.0 [40.0,60.0] | 50.0 [40.0,60.0] | 50.0 [40.0,80.0] | <0.001 |
| PEEP, median [Q1,Q3] | 5.0 [5.0,8.0] | 5.0 [5.0,8.0] | 5.0 [5.0,10.0] | <0.001 |
| Tidal Volume, median [Q1,Q3] | 522.0 [450.0,608.0] | 525.0 [450.0,610.5] | 506.0 [440.0,600.0] | <0.001 |
| UrineOutput_IO, median [Q1,Q3] | 80.0 [40.0,150.0] | 80.0 [40.0,160.0] | 50.0 [22.0,100.0] | <0.001 |
| PaO2/FiO2, median [Q1,Q3] | 2.6 [1.7,3.7] | 2.6 [1.8,3.7] | 2.1 [1.3,3.5] | <0.001 |
| SIRS, median [Q1,Q3] | 1.0 [0.0,2.0] | 1.0 [0.0,2.0] | 2.0 [1.0,3.0] | <0.001 |
| Shock_Index, median [Q1,Q3] | 0.7 [0.6,0.9] | 0.7 [0.6,0.8] | 0.8 [0.7,1.0] | <0.001 |
| SOFA, median [Q1,Q3] | 2.0 [0.0,5.0] | 2.0 [0.0,4.0] | 5.0 [3.0,8.0] | <0.001 |
| SAPSII, median [Q1,Q3] | 29.0 [20.0,37.0] | 28.0 [19.0,36.0] | 41.0 [30.0,53.0] | <0.001 |
| OASIS, median [Q1,Q3] | 19.0 [16.0,22.0] | 19.0 [16.0,21.0] | 22.0 [19.0,29.0] | <0.001 |
| GCS Total, median [Q1,Q3] | 15.0 [10.0,15.0] | 15.0 [10.0,15.0] | 8.0 [5.0,14.0] | <0.001 |

**Supplementary Table 4:** *Distribution of Data Points for MIMIC-IV*

| | Overall | Survived | Dead | P-Value |
|---|---|---|---|---|
| ICU-Stays | 60087 | 56299 | 3788 | |
| age, median [Q1,Q3] | 67.0 [55.0,77.0] | 66.0 [55.0,77.0] | 72.0 [60.0,82.0] | <0.001 |
| gender, n (%) - Male | 33913 (56.4) | 31850 (56.6) | 2063 (54.5) | 0.012 |
| gender, n (%) - Female | 26174 (43.6) | 24449 (43.4) | 1725 (45.5) | |
| race, n (%) - Asian | 1747 (2.9) | 1628 (2.9) | 119 (3.1) | <0.001 |
| race, n (%) - Black/African American | 6300 (10.5) | 5948 (10.6) | 352 (9.3) | |
| race, n (%) - Hispanic/Latino | 2277 (3.8) | 2172 (3.9) | 105 (2.8) | |
| race, n (%) - Native American | 114 (0.2) | 107 (0.2) | 7 (0.2) | |
| race, n (%) - Other | 2096 (3.5) | 1983 (3.5) | 113 (3.0) | |
| race, n (%) - Unknown | 6351 (10.6) | 5614 (10.0) | 737 (19.5) | |
| race, n (%) - White | 41202 (68.6) | 38847 (69.0) | 2355 (62.2) | |
| Ethnicity, n (%) - Hispanic | 2277 (3.8) | 2172 (3.9) | 105 (2.8) | 0.001 |
| Ethnicity, n (%) - Non Hispanic | 57810 (96.2) | 54127 (96.1) | 3683 (97.2) | |
| icuLos_h, median [Q1,Q3] | 53.6 [33.2,99.5] | 52.3 [32.7,95.5] | 98.5 [47.3,206.4] | <0.001 |
| Number of Notes, median [Q1,Q3] | 2.0 [1.0,2.0] | 2.0 [1.0,2.0] | 2.0 [1.0,3.0] | <0.001 |
| Heart Rate, median [Q1,Q3] | 83.0 [72.0,96.0] | 83.0 [72.0,96.0] | 90.0 [76.0,105.0] | <0.001 |
| SpO2, median [Q1,Q3] | 97.0 [95.0,99.0] | 97.0 [95.0,99.0] | 97.5 [95.0,100.0] | <0.001 |
| Oxygen Saturation, median [Q1,Q3] | 95.0 [76.0,98.0] | 95.0 [77.0,98.0] | 92.0 [72.5,97.0] | <0.001 |
| Respiratory Rate, median [Q1,Q3] | 18.5 [15.5,22.0] | 18.0 [15.0,22.0] | 21.0 [17.0,25.0] | <0.001 |
| Temperature, median [Q1,Q3] | 36.8 [36.6,37.2] | 36.8 [36.6,37.2] | 36.8 [36.3,37.3] | <0.001 |
| Non Invasive Blood Pressure mean, median [Q1,Q3] | 76.0 [67.0,87.0] | 76.5 [67.0,87.0] | 72.0 [63.0,82.0] | <0.001 |
| Non Invasive Blood Pressure systolic, median [Q1,Q3] | 116.0 [102.5,132.0] | 116.0 [103.0,132.0] | 109.0 [96.0,124.0] | <0.001 |
| Non Invasive Blood Pressure diastolic, median [Q1,Q3] | 63.0 [54.0,74.0] | 63.0 [54.0,74.0] | 60.0 [51.0,70.0] | <0.001 |
| Glucose, median [Q1,Q3] | 131.0 [108.0,166.0] | 131.0 [108.0,164.0] | 142.0 [108.0,196.0] | <0.001 |
| Creatinine, median [Q1,Q3] | 1.0 [0.7,1.6] | 1.0 [0.7,1.5] | 1.6 [1.0,2.6] | <0.001 |
| Base Excess, median [Q1,Q3] | -1.0 [-4.0,0.5] | -1.0 [-3.0,1.0] | -4.0 [-9.0,0.0] | <0.001 |
| BUN, median [Q1,Q3] | 20.0 [13.0,35.0] | 19.0 [13.0,33.0] | 32.0 [20.0,52.0] | <0.001 |
| Anion Gap, median [Q1,Q3] | 14.0 [12.0,17.0] | 14.0 [12.0,16.0] | 17.0 [14.0,21.0] | <0.001 |
| Bicarbonate, median [Q1,Q3] | 23.0 [20.0,26.0] | 23.0 [20.0,26.0] | 20.0 [17.0,24.0] | <0.001 |
| Lactate, median [Q1,Q3] | 2.0 [1.4,3.2] | 1.9 [1.3,2.9] | 3.3 [1.9,6.2] | <0.001 |
| Hemoglobin, median [Q1,Q3] | 10.0 [8.7,11.5] | 10.0 [8.7,11.5] | 9.6 [8.3,11.2] | <0.001 |
| Hematocrit, median [Q1,Q3] | 30.1 [26.3,34.4] | 30.1 [26.3,34.5] | 29.5 [25.6,34.3] | <0.001 |
| pH, median [Q1,Q3] | 7.4 [7.3,7.4] | 7.4 [7.3,7.4] | 7.3 [7.2,7.4] | <0.001 |
| Bilirubin, Direct, median [Q1,Q3] | 0.4 [0.1,1.1] | 0.4 [0.1,1.1] | 0.6 [0.2,2.1] | <0.001 |
| pO2, median [Q1,Q3] | 118.0 [80.0,195.0] | 122.0 [82.0,202.0] | 95.0 [66.0,142.0] | <0.001 |
| pCO2, median [Q1,Q3] | 40.0 [35.5,46.0] | 40.0 [36.0,46.0] | 39.0 [33.0,47.0] | <0.001 |
| AST, median [Q1,Q3] | 45.0 [25.0,109.0] | 42.0 [24.0,99.0] | 79.0 [36.0,211.0] | <0.001 |
| ALT, median [Q1,Q3] | 31.0 [17.0,78.0] | 30.0 [17.0,72.0] | 45.0 [21.0,135.0] | <0.001 |
| WBC, median [Q1,Q3] | 10.8 [7.4,15.1] | 10.7 [7.4,14.9] | 12.6 [8.0,18.8] | <0.001 |
| RBC, median [Q1,Q3] | 5.0 [1.0,18.0] | 4.0 [1.0,17.0] | 8.0 [2.0,27.0] | <0.001 |
| Potassium, median [Q1,Q3] | 4.2 [3.8,4.6] | 4.2 [3.8,4.6] | 4.2 [3.8,4.8] | <0.001 |
| Sodium, median [Q1,Q3] | 138.0 [135.0,141.0] | 138.0 [135.0,141.0] | 138.0 [134.5,142.0] | <0.001 |
| Chloride, median [Q1,Q3] | 105.0 [101.0,108.0] | 105.0 [101.0,108.0] | 104.0 [99.0,109.0] | <0.001 |
| Magnesium, median [Q1,Q3] | 2.0 [1.8,2.3] | 2.0 [1.8,2.3] | 2.1 [1.8,2.3] | <0.001 |
| Phosphate, median [Q1,Q3] | 3.5 [2.8,4.4] | 3.5 [2.8,4.3] | 4.2 [3.2,5.6] | <0.001 |
| FiO2, median [Q1,Q3] | 50.0 [40.0,60.0] | 50.0 [40.0,60.0] | 50.0 [40.0,70.0] | <0.001 |
| PEEP, median [Q1,Q3] | 5.2 [5.0,9.0] | 5.0 [5.0,8.0] | 7.0 [5.0,10.5] | <0.001 |
| Tidal Volume, median [Q1,Q3] | 469.0 [402.0,536.0] | 472.0 [405.0,539.5] | 450.0 [392.0,514.0] | <0.001 |
| UrineOutput_IO, median [Q1,Q3] | 80.0 [40.0,170.0] | 80.0 [40.0,175.0] | 45.0 [20.0,100.0] | <0.001 |
| PaO2/FiO2, median [Q1,Q3] | 2.5 [1.6,3.6] | 2.5 [1.6,3.6] | 1.9 [1.1,3.2] | <0.001 |
| SIRS, median [Q1,Q3] | 1.0 [0.0,2.0] | 1.0 [0.0,2.0] | 2.0 [1.0,3.0] | <0.001 |
| Shock_Index, median [Q1,Q3] | 0.7 [0.6,0.9] | 0.7 [0.6,0.8] | 0.8 [0.7,1.0] | <0.001 |
| SOFA, median [Q1,Q3] | 3.0 [0.0,6.0] | 3.0 [0.0,6.0] | 8.0 [4.0,10.0] | <0.001 |
| SAPSII, median [Q1,Q3] | 29.0 [22.0,40.0] | 29.0 [21.0,39.0] | 46.0 [33.0,59.0] | <0.001 |
| OASIS, median [Q1,Q3] | 19.0 [16.0,22.0] | 19.0 [16.0,21.0] | 23.0 [19.0,29.0] | <0.001 |
| Total GCS, median [Q1,Q3] | 15.0 [10.0,15.0] | 15.0 [11.0,15.0] | 8.0 [5.0,14.0] | <0.001 |

**Supplement Text 1:** *Variable Definitions and Data Harmonization*

Below is a detailed description of all the data types used in the predictive models:

**Time-invariant (Static) EHR data.** Time-invariant data included patient demographics and baseline comorbidities. If not directly available, age was calculated based on the patient's date of birth and admission date. Patient gender was converted into a binary representation, and race was encoded into numerical format by assigning a number to each race/ethnicity category. Finally, we extracted comorbidities for patients admitted to the ICU, utilizing both ICD-9 and ICD-10 coding systems.[1] We identified a set of 30 specific comorbid conditions, including congestive heart failure, hypertension, cardiac arrhythmias, diabetes, and liver disease, among others. Each patient's data was transformed into a binary vector of length 30, indicating the presence (1) or absence (0) of each comorbidity based on their diagnostic codes. Codes were selected based on their direct relevance to the defined comorbidities and were processed as a binary vector format that facilitates straightforward integration with the multimodal models.

**Time-variant (Dynamic) EHR data.** The time-variant features included laboratory values, medications, ventilator settings, clinical scores (e.g., clinical risk scores, Glasgow Coma Scale [GCS], etc.), clinical events (e.g., procedures), and hemodynamics. The extraction and preprocessing of dynamic EHR data began with cleaning each data table separately. This involved removing missing measurements and dropping rare events. Numeric and textual variables were standardized as numbers and string types, while categorical variables were encoded into numeric representations. Specifically, string variables initially recorded numerically, such as ventilation mode, were converted to descriptive string names for consistency. Also, numerical variables that were erroneously recorded as strings in the database were converted to float types. Additionally, variable names were standardized when the same variable appears in data fields across different tables to ensure consistency. Variable units were also converted as necessary to guarantee comparability between different data sources, ensuring that all variables were represented with consistent names and measurement units across tables. Outliers in any dataset were detected and removed based on a clinically valid range for each variable, guided by clinical knowledge for these decisions. Dynamic variables were binned into non-overlapping 1-hour windows, within which the mean for continuous variables or the mode for categorical variables were calculated when multiple values were recorded within a bin. This approach allowed us to accommodate for varying sampling frequencies across measurements.

To harmonize the medication tables, we utilized regular expressions to identify and extract key medications, and subsequently standardized their names. We created extensive lists of drug names for each critical care medication to ensure thorough coverage during the extraction process. Then, we converted the presence of these drugs into a binary representation that indicated whether patients received them or not. Vasopressor use was defined based on whether a patient received any of the following medications: epinephrine, vasopressin, milrinone, dobutamine, phenylephrine, or norepinephrine. To identify antibiotic medications, specific criteria were applied. Initially, a subset of data focusing on active antibiotic agents was created. Further refinement included removing entries based on the administration route; drugs applied to the eyes or ears and those identified with specific routes like 'OU' (*oculus uterque*, i.e., both eyes) were excluded, in order to focus the analyses on parenterally administered antibiotics. Then, utilizing a list of known antibiotic substances, including both common names and specific drug identifiers (GSN codes), any drug's name that contained any part of the antibiotic substrings or matched any GSN codes was flagged as an antibiotic with a binary indicator set to one.[2] A similar methodical approach to handling and standardizing urine output measurements in clinical datasets was essential for consistent data analysis where accurate fluid balance assessments are necessary. The process began with the creation of a list that included various descriptors and terms used to denote urine output within the dataset. This list encompassed a range of expressions, from general terms like 'Urine' and 'URINE CATHETER' to more specific descriptors such as 'Output External Urethral Device Condom Catheter' and 'Urine Output-Foley.' This inclusive strategy ensured that all potential variations in urine output recording were captured.

To manage missing values in the datasets, we implemented a four-step strategy. First, we imputed missing data by estimating values based on their relationships with other measured variables. For example, to impute missing Glasgow Coma Scale (GCS) total scores, we used the available motor, verbal, and eye component scores. Additionally, we employed a method to estimate GCS totals using the Richmond Agitation-Sedation Scale (RASS), which assessed patient agitation and sedation levels. We established a mapping based on clinical correlations between RASS and GCS scores, where a RASS score of 0 (alert and calm) corresponded to a full GCS total score of 15 (indicating full consciousness). For Hemoglobin (Hb) and Hematocrit (Hct), we leveraged their linear relationship to impute missing values when only one measurement was available. In cases of bilirubin levels, we established ratios to estimate either total or direct bilirubin based on the availability of one type. For arterial blood pressures, we used the relationships among systolic, diastolic, and mean arterial pressures to estimate missing values if one of the three variables was missing. Second, we introduced a binary missing value indicator for each time step, which identified whether a measurement is actual or imputed.[3] Third, we employed the "Sample and Hold" strategy for imputation, maintaining imputed values for up to 24 hours for medications and 12 hours for other dynamic variables.[4] Finally, to prevent data leakage, we addressed the rest of missing values in our dataset post-split by imputing them using the mean, mode, and zero for continuous, categorical, and binary variables respectively, based on the training data. Subsequently, the continuous variables are normalized to have a zero mean and unit variance. All time-variant patient data sequences were truncated and padded to encapsulate only the first 24 hours of the patients' stay.

**Time-series Vital Signs.** The extraction and preprocessing of vital signs started by standardization of variable names across different tables. Outliers were removed using clinically valid interval ranges for each variable. However, unlike the MIMIC dataset (as vital sign measurements were no more frequent than per hour), vital sign measurements in the HiRID and eICU databases were calculated using non-overlapping 5-minute windows, which provided more granular tracking of patient conditions over time.

To deal with the high frequency measurements of vital signs in ICU settings, we leveraged two approaches: using raw data recorded every minute or aggregating measurements into hourly averages over non-overlapping windows. The raw data approach preserved high temporal resolution and captured fine-grained changes in a patient's physiological state and short-term fluctuations, but it may introduce noise, increase computational demands, and risk overfitting due to the voluminous data. Conversely, hourly averaging simplifies the data, reduces noise and computational requirements but might lose valuable temporal information and smooth out critical short-term fluctuations essential for accurate predictions. To address these challenges, we employed a vital sign frequency normalization pipeline by extending the feature engineering methods provided by the open-source POBM toolbox, and adapting its $SPO_2$-focused approach to other vital signs available in the ICU.[5,6]

Our pipeline can be divided into seven categories: 1) General Statistics: Metrics describing variable distribution and behavior over a given period, such as range, percentiles, below median percentage, and zero crossings (signal crossing a baseline or mean value).[7] 2) Complexity: Metrics assessing the complexity of time-series variables, indicative of long-term fluctuations and correlations. Approximate entropy (ApEn) and sample entropy (SampEn) are commonly used as measures of unpredictability.[8,9] 3) Periodicity: Metrics identifying and quantifying repeated or cyclical patterns, such as autocorrelation, which measures the correlation between values of the same variable over time. 4) Abnormal Patterns: Descriptive statistics of drops and elevations in variables, providing information on their severity, duration, and frequency. Examples include desaturation in $SpO_2$ (computed as area, slope, length and depth of the desaturations), bradycardia and tachycardia in heart rate (HR), and tachypnea and bradypnea in respiratory rate (RR). 5) Stress Burden: Metrics quantifying the extent and impact of abnormal vital sign levels on the body, such as hypoxic burden (e.g., computed as total time $SpO_2$ is below a certain threshold), cardiac stress burden, and respiratory stress burden. 6) Fourier Transform: We compute key spectral features using the Fast Fourier Transform (FFT), which converts time-domain signals into their frequency-domain representation. This allows us to extract features such as peak frequency (the dominant frequency component), spectral centroid (the center of mass of the spectrum), spectral bandwidth (the spread of frequencies), spectral flatness (the tonality vs. noisiness of the signal), spectral energy (total power of the signal), and spectral entropy (complexity of the spectral distribution). 7) Wavelet Transform: We employ a wavelet-based feature extraction methodology using the Discrete Wavelet Transform (DWT) with the Daubechies 4 (db4) wavelet to capture both time and frequency domain characteristics of the signals. Each signal is decomposed up to level 2, allowing us to analyze both low-frequency (approximation coefficients) and high-frequency (detail coefficients) components at multiple scales. From these coefficients, we computed statistical features such as mean, variance, and standard deviation.

**Clinical Notes.** First, regular expressions were utilized to identify and mask protected health information (PHI) placeholders, represented by [* *] in the MIMIC database, by replacing them with spaces. Furthermore, the text was truncated beyond predefined keywords that signal the formal conclusion of documents, such as electronic signatures or terms commonly found at the end of clinical reports. Then, we implemented a method to structure and filter clinical text notes by identifying specific section headings, such as 'EXAMINATION', 'FINDINGS', 'HISTORY', 'IMPRESSION', and 'INDICATION', using a comprehensive regular expression pattern.[10] Subsequently, matches to these headings were iteratively searched, and the content was extracted. The extracted content was then checked to ensure it contained text rather than just whitespace or non-informative markers. Valid content was normalized by condensing excessive whitespace, converting accented characters to their ASCII equivalents, and removing punctuation using a predefined set of punctuation characters. Next, we compiled a comprehensive list of ~1500 abbreviations, each with at least three characters. We then constructed a regular expression pattern to recognize these abbreviations as distinct words within the text, while preventing incorrect expansions within longer word forms. Finally, sentences with less than 10 words were removed because they usually lack sufficient context to be useful for our analysis.

To prepare clinical texts for language model training, we initially used a tokenizer to truncate the input text to a maximum allowable length of 512 tokens, while also generating any overflowing tokens. This ensures that longer texts are not discarded but split into several contiguous sequences. However, overflow sequences containing less than 75 tokens were excluded.

**Chest X-ray Images.** Only MIMIC-IV had chest x-ray data. In our study, preprocessing of chest X-ray images involved resizing and cropping to 224x224 pixels while maintaining aspect ratios with zero-padding. Bilateral filtering[11] was applied for noise reduction while preserving edge details, and gamma correction[12] was used to enhance visibility of details. Furthermore, contrast was enhanced using Contrast-Limited Adaptive Histogram Equalization (CLAHE),[13] which segments images into non-overlapping regions and applies histogram equalization with a clipping and redistribution procedure to limit noise and prevent over-exposure.

We used five augmentation techniques on-the-fly while loading images for training, including random horizontal flipping with a 50% probability, random rotations up to 10 degrees, and random resized cropping with scales between 80% to 100%. Additionally, we increased sharpness by a factor of 1.5 in 50% of the images. We also adjusted brightness, contrast, saturation, and hue slightly, applying these changes in 10% of the cases. Finally, the images were converted to tensors and normalized.

**Supplement Figure 1**. *Data harmonization and normalization steps conducted for each data type prior to model development*: 1) time-invariant data: demographics and baseline comorbidities; 2) clinical notes, 3) time-variant electronic health record data, including labs, procedures, medications, risk score, ventilator settings, and fluid intake/output; 4) time-variant vital signs data; and 5) chest X-ray imaging.

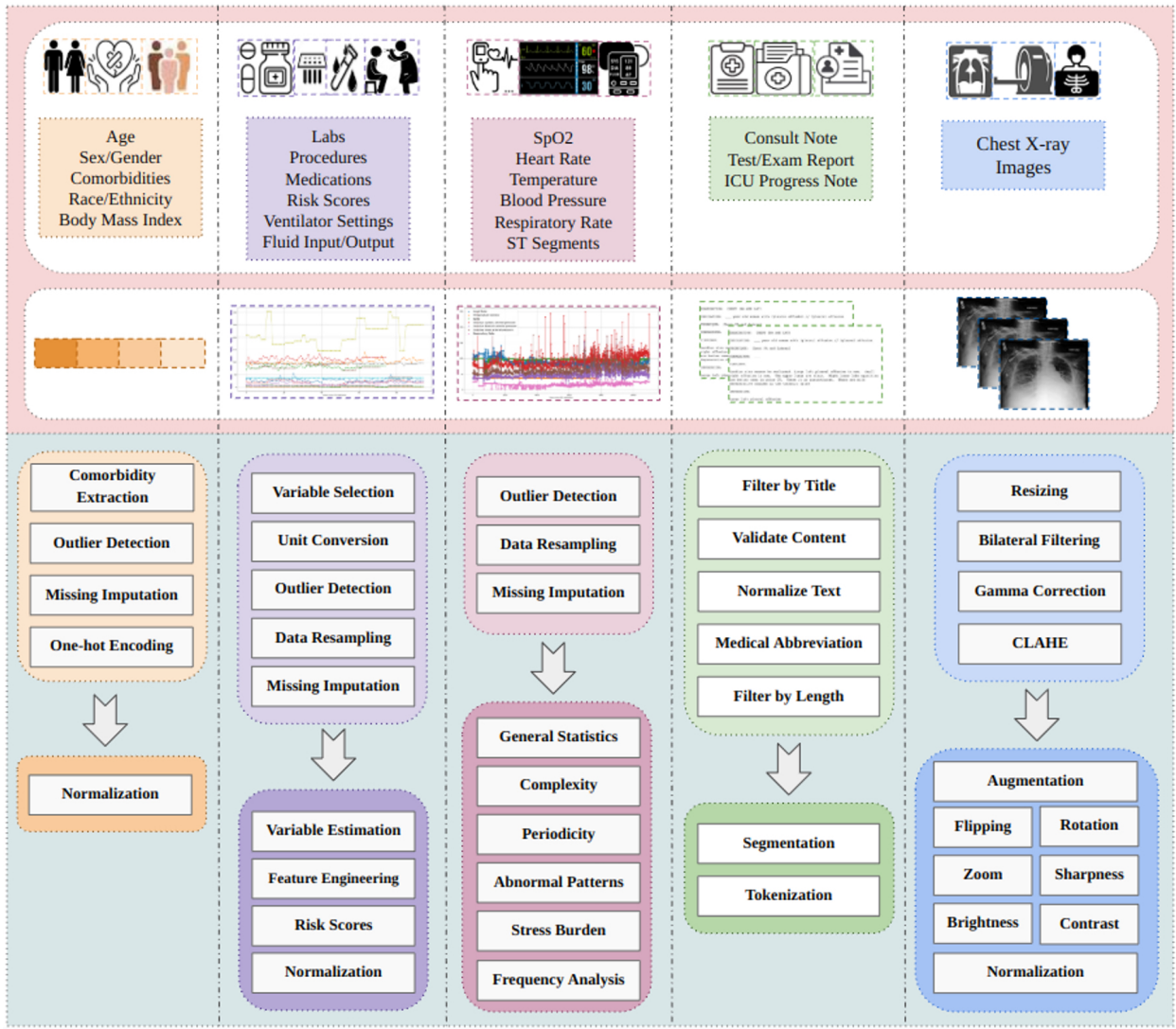

**Supplement Text 2:** *Model Evaluation Setting*

*Study Design*
We present findings from our assessment of the generalizability performance of our prediction models, structured around four incremental steps. First, we assessed the performance of models using only vital signs recorded at high frequency and in combination with the other time-variant EHR data as baseline data. Second, we evaluated models that utilize time-invariant EHR data, both as standalone models and when integrated with the time-variant EHR data. Third, we explored the effectiveness of models that incorporate clinical notes, assessing their predictive capabilities alone and in combination with other data modalities. Fourth, we examine models employing CXR independently and within an extensive multimodal framework incorporating all previously mentioned modalities.

*Internal Evaluation Setting*
In our multimodal learning problems, it was essential to ensure that each data split—training, validation, and test sets—accurately represented the overall dataset in terms of both mortality outcomes and modality availability. Therefore, we first calculated the mortality rate across the entire dataset to maintain this rate across all splits using stratified sampling. Recognizing that not all patients have all data modalities, we created subgroups based on the availability of data modalities. Within each subgroup, we used stratification to split patients into the train, validation, and test sets. This stratification approach ensured that each subset accurately reflected the full dataset's modality and mortality distribution, which may effectively minimize outcome bias during model evaluation. For training and evaluating our predictive models, we split the data into 70% for training, 10% for validation, and 20% for testing.

*External Validation Setting*
We performed external validation of the models developed in the first and second steps to assess their generalizability. For the models based on baseline data combined with high-frequency vital signs in the first step, we created a test set from the eICU multi-center database by selecting patients from eight unique hospitals, each with more than 2,500 ICU stays. We ensured no patients from these eight hospitals were included in the training or validation datasets. Then, for the models based on baseline data combined with hourly measured vital signs and time-invariant EHR data in the second step, we developed models using patients admitted before 2019 from the MIMIC datasets and externally validated using patients admitted during the COVID-19 period (2020–2022), as well as patients from HiRID, eICU, and the eight unique hospitals. The patient data collected during the COVID-19 pandemic are considered "external" relative to pre-pandemic data from the same institution due to the significant shifts in ICU care protocols, resource allocation, and patient demographics that occurred during this period[14,15].

*Performance Evaluation*
We considered several performance metrics to evaluate the predictive performance of our models. The performance metrics employed included area under the receiver operating characteristic curve (AUROC), and area under the precision-recall curve (AUPRC), F1-score, precision, specificity, the Matthews correlation coefficient (MCC)[16], and brier score[17]. The decision threshold for classifying model output probability was chosen according to an 80% sensitivity level.

*Fairness and Bias*

To ensure fairness of our models across various social cohorts, we conducted a bias audit using the Aequitas toolkit - an open-source library for bias and fairness analysis[18]. Aequitas toolkit assesses fairness of our model across gender, age, and race by examining True Positive Rate (TPR), True Negative Rate (TNR), False Positive Rate (FPR), and False Negative Rate (FNR) across different patient subgroups to identify any significant deviations from established fairness thresholds.

*Model Transparency and Interpretability*

We adopted the Integrated Gradients (IG) method[19] to address the interpretability of our Bi-LSTM and BERT models used in analyzing dynamic EHR data and clinical notes, respectively. IG identifies influential parts of the input data by comparing the model's prediction for the actual input with a baseline input— a sequence of zeros or padding tokens of the same length as the input. This method involves interpolating between the baseline and actual input, then integrating the gradients of the model's output with respect to the input along with this interpolation path. The result is a set of attributions that highlight which aspects of the input are the most responsible for the model's prediction. For EHR data, attributions for each feature are aggregated across all time steps to identify key influences on the model's prediction. Positive attributions increase the prediction likelihood for a specific class, thereby supporting the model's decision, while negative attributions decrease this likelihood. For clinical notes, we generated our interpretation of the BERT model using the LayerIntegratedGradients method from the Captum library[20], which attributes predictions to individual neurons and quantifies their contributions across layers. We focused on the embedding layer to understand how input tokens affect model predictions, processing raw attributions to determine each token's impact. We employed a color-coded visualization scheme, with red indicating positive influences on the model's prediction by increasing the risk of mortality, and green indicating negative influences, decreasing the risk of mortality. For interpretation of the DenseNet121 model used to analyze CXR, we applied Gradient-weighted Class Activation Mapping (Grad-CAM)[21,22]. Grad-CAM highlights the spatial regions that are the most influential in making predictions as heatmaps.

**Supplement Figure 2.** *Histogram illustrating the distribution of time from ICU admission to death (days) for each dataset.*

## MIMIC-III

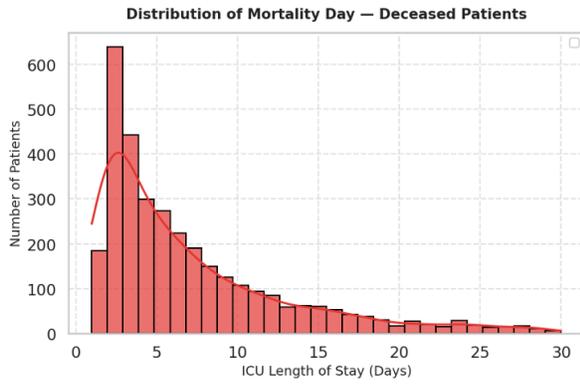

## MIMIC-IV

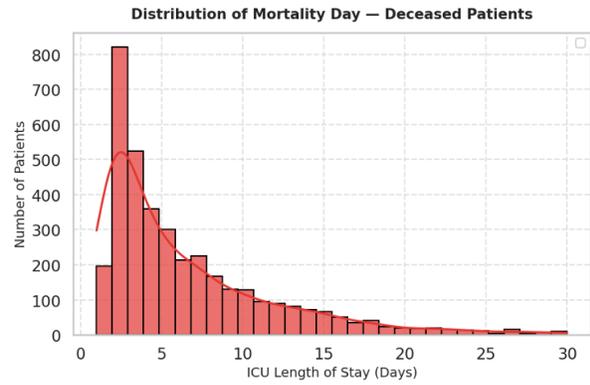

## eICU

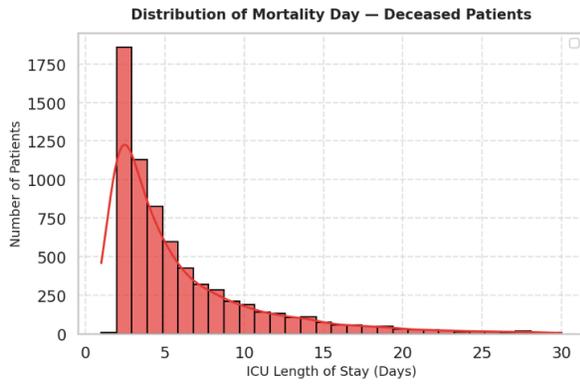

## HiRID

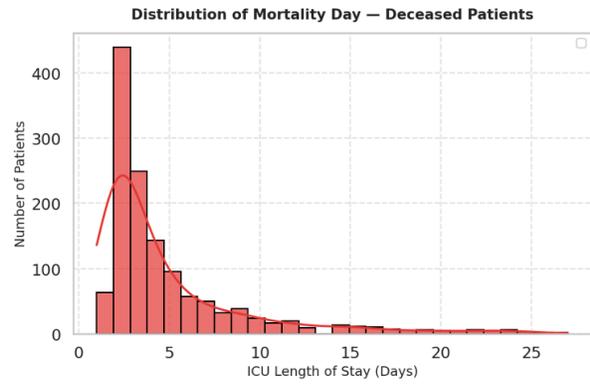

**Supplement Figure 5.** *Fairness/bias analyses for MIMIC datasets.* We calculated the true positive (TPR), true negative (TNR), false positive (FPR), and false negative (FNR) rates when the model was tested on specific cohorts based on race/ethnicity, age group, or sex on data from a) MIMIC-III, and b) MIMIC-IV.

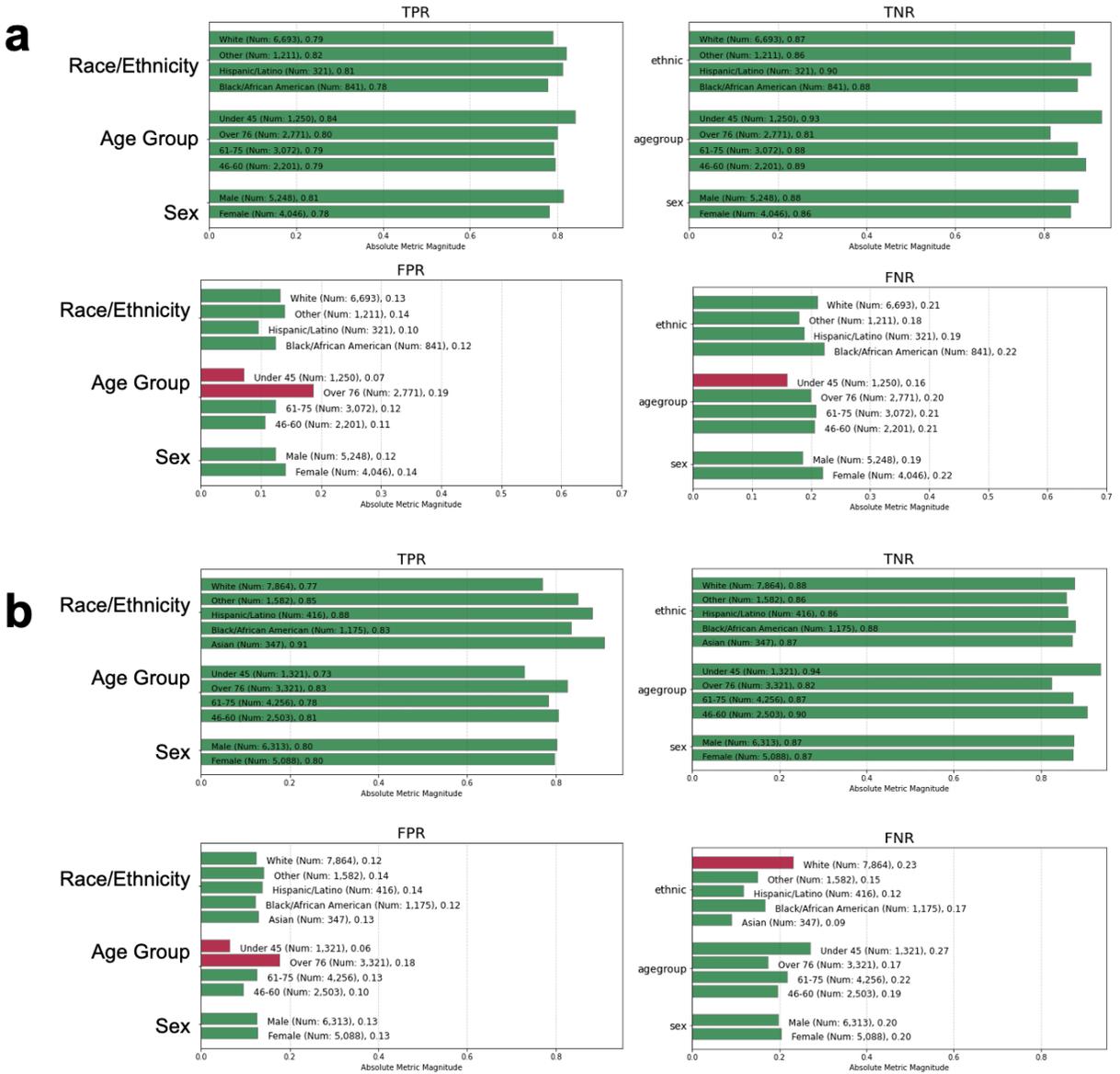

**Supplement Table 5.** *Performance metrics for predicting mortality using time-variant and time-invariant structured EHR data on MIMIC-III and MIMIC-IV data.*

**MIMIC-III:** The analyzed cohort comprised 46,983 ICU stays, including 26,603 (56.6%) male and 20,380 (43.4%) female patients. The overall mortality rate observed was 7%. The median age of the patients was 66 years, with an interquartile range (IQR) of 53 to 78 years.

| MIMIC-III | | | | | | | | |
|---|---|---|---|---|---|---|---|---|
| Model | ACC (95% CI) | PRC (95% CI) | F-1 (95% CI) | SPE (95% CI) | AUC (95% CI) | AP (95% CI) | MCC (95% CI) | Brier (95% CI) |
| **Time-Invariant** | 0.520 (0.509-0.531) | 0.110 (0.102-0.119) | 0.193 (0.180-0.207) | 0.498 (0.487-0.509) | 0.717 (0.697-0.737) | 0.178 (0.156-0.204) | 0.154 (0.136-0.171) | 0.145 (0.143-0.147) |
| **Time-Variant Baseline** | 0.841 (0.834-0.848) | 0.285 (0.264-0.304) | 0.420 (0.395-0.443) | 0.844 (0.836-0.851) | 0.906 (0.895-0.915) | 0.471 (0.429-0.513) | 0.414 (0.389-0.437) | 0.261 (0.255-0.267) |
| **Combined Modalities** | 0.864 (0.857-0.871) | 0.321 (0.299-0.342) | 0.458 (0.434-0.483) | 0.869 (0.862-0.876) | 0.916 (0.904-0.925) | 0.531 (0.491-0.568) | 0.451 (0.426-0.475) | 0.191 (0.185-0.197) |

**MIMIC-IV:** The analyzed cohort included 60,087 ICU stays, comprising 33,913 (56.4%) male and 26,174 (43.6%) female patients. The observed mortality rate was 6%. Patients had a median age of 67 years, with an interquartile range (IQR) of 55 to 77 years.

| MIMIC-IV | | | | | | | | |
|---|---|---|---|---|---|---|---|---|
| Model | ACC (95% CI) | PRC (95% CI) | F-1 (95% CI) | SPE (95% CI) | AUC (95% CI) | AP (95% CI) | MCC (95% CI) | Brier (95% CI) |
| **Time-Invariant** | 0.553 (0.543-0.561) | 0.106 (0.098-0.115) | 0.187 (0.175-0.201) | 0.535 (0.526-0.545) | 0.736 (0.719-0.755) | 0.181 (0.158-0.206) | 0.165 (0.149-0.181) | 0.389 (0.385-0.393) |
| **Time-Variant Baseline** | 0.853 (0.847-0.860) | 0.279 (0.260-0.300) | 0.413 (0.390-0.438) | 0.857 (0.850-0.864) | 0.912 (0.903-0.921) | 0.473 (0.434-0.512) | 0.416 (0.394-0.439) | 0.282 (0.277-0.289) |
| **Combined Modalities** | 0.869 (0.863-0.875) | 0.305 (0.283-0.325) | 0.440 (0.417-0.464) | 0.874 (0.867-0.879) | 0.921 (0.912-0.930) | 0.518 (0.476-0.558) | 0.441 (0.418-0.464) | 0.064 (0.061-0.067) |

Abbreviations: CI, confidence interval; ACC, Accuracy; PRC, Precision; SPE, Specificity; AUC, Area under the Receiver Operating Characteristics Curve; AP, Area under the Precision-Recall Curve; MCC, Matthews Correlation Coefficient.

**Supplemental Figure 6**. *Fairness/bias analyses for the HiRID and eICU datasets.* We calculated the true positive (TPR), true negative (TNR), false positive (FPR), and false negative (FNR) rates when the model was tested on specific cohorts based on race/ethnicity, age group, or sex on data from a) HiRID, and b) eICU.

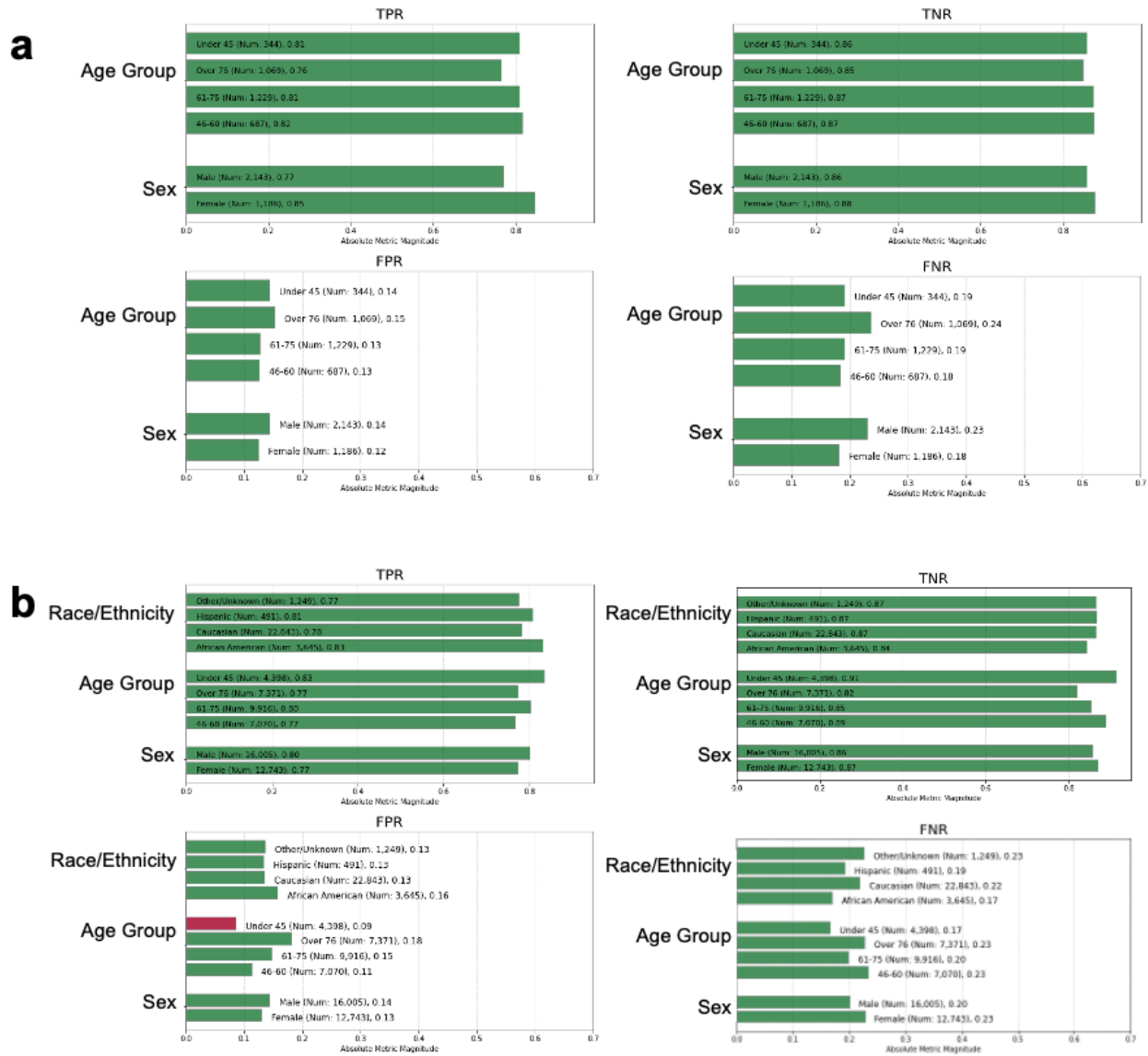

**Supplement Table 6.** Performance metrics on external validation on MIMIC-IV (Covid era), HiRID, eICU, and 8 separate hospitals from eICU separately.

| Metric/ Dataset | ACC (95% CI) | PRC (95% CI) | F-1 (95% CI) | SPE (95% CI) | AUC (95% CI) | AP (95% CI) | MCC (95% CI) | Brier (95% CI) |
|---|---|---|---|---|---|---|---|---|
| **External Validation** | | | | | | | | |
| **Covid era (MIMIC-IV)** | 0.873 (0.866-0.880) | 0.369 (0.343-0.393) | 0.504 (0.477-0.530) | 0.879 (0.872-0.886) | 0.923 (0.915-0.932) | 0.539 (0.496-0.577) | 0.487 (0.458-0.512) | 0.194 (0.188-0.201) |
| **HiRID** | 0.829 (0.823-0.834) | 0.290 (0.276-0.304) | 0.426 (0.409-0.443) | 0.831 (0.825-0.836) | 0.890 (0.881-0.899) | 0.480 (0.450-0.510) | 0.413 (0.396-0.429) | 0.203 (0.200-0.207) |
| **eICU** | 0.770 (0.768-0.772) | 0.159 (0.156-0.163) | 0.266 (0.260-0.272) | 0.768 (0.766-0.770) | 0.865 (0.861-0.870) | 0.344 (0.333-0.356) | 0.287 (0.282-0.293) | 0.178 (0.177-0.179) |
| **Hospital-1** | 0.786 (0.772-0.800) | 0.163 (0.138-0.189) | 0.270 (0.231-0.305) | 0.787 (0.772-0.801) | 0.841 (0.808-0.872) | 0.307 (0.235-0.383) | 0.288 (0.247-0.325) | 0.168 (0.160-0.176) |
| **Hospital-2** | 0.715 (0.696-0.732) | 0.210 (0.182-0.238) | 0.338 (0.299-0.374) | 0.701 (0.682-0.719) | 0.850 (0.823-0.875) | 0.406 (0.341-0.473) | 0.328 (0.292-0.364) | 0.214 (0.204-0.225) |
| **Hospital-3** | 0.740 (0.722-0.758) | 0.193 (0.167-0.220) | 0.313 (0.277-0.350) | 0.733 (0.714-0.751) | 0.871 (0.844-0.895) | 0.413 (0.338-0.486) | 0.315 (0.275-0.351) | 0.198 (0.187-0.208) |
| **Hospital-4** | 0.807 (0.795-0.820) | 0.283 (0.253-0.310) | 0.415 (0.379-0.447) | 0.811 (0.797-0.824) | 0.879 (0.861-0.897) | 0.498 (0.439-0.556) | 0.388 (0.351-0.423) | 0.145 (0.138-0.152) |
| **Hospital-5** | 0.759 (0.743-0.776) | 0.192 (0.164-0.220) | 0.312 (0.273-0.350) | 0.753 (0.737-0.771) | 0.886 (0.861-0.909) | 0.449 (0.373-0.524) | 0.325 (0.286-0.361) | 0.184 (0.174-0.192) |
| **Hospital-6** | 0.725 (0.709-0.741) | 0.191 (0.167-0.217) | 0.315 (0.281-0.351) | 0.712 (0.694-0.728) | 0.894 (0.872-0.915) | 0.398 (0.332-0.465) | 0.333 (0.300-0.365) | 0.209 (0.199-0.219) |
| **Hospital-7** | 0.726 (0.709-0.744) | 0.158 (0.133-0.185) | 0.268 (0.231-0.309) | 0.716 (0.698-0.734) | 0.896 (0.875-0.917) | 0.396 (0.315-0.483) | 0.304 (0.270-0.341) | 0.205 (0.195-0.214) |
| **Hospital-8** | 0.803 (0.788-0.818) | 0.217 (0.188-0.249) | 0.347 (0.307-0.389) | 0.799 (0.783-0.814) | 0.921 (0.902-0.938) | 0.474 (0.396-0.555) | 0.372 (0.336-0.409) | 0.155 (0.146-0.162) |

Abbreviations: CI, confidence interval; ACC, Accuracy; PRC, Precision; SPE, Specificity; AUC, Area under the Receiver Operating Characteristics Curve; AP, Area under the Precision-Recall Curve; MCC, Matthews Correlation Coefficient.

**Supplement Figure 7:** *Variable importance analysis of structured data for a) HiRID and b) eICU datasets*

**a** *Variable importance - external validation on HiRID dataset*

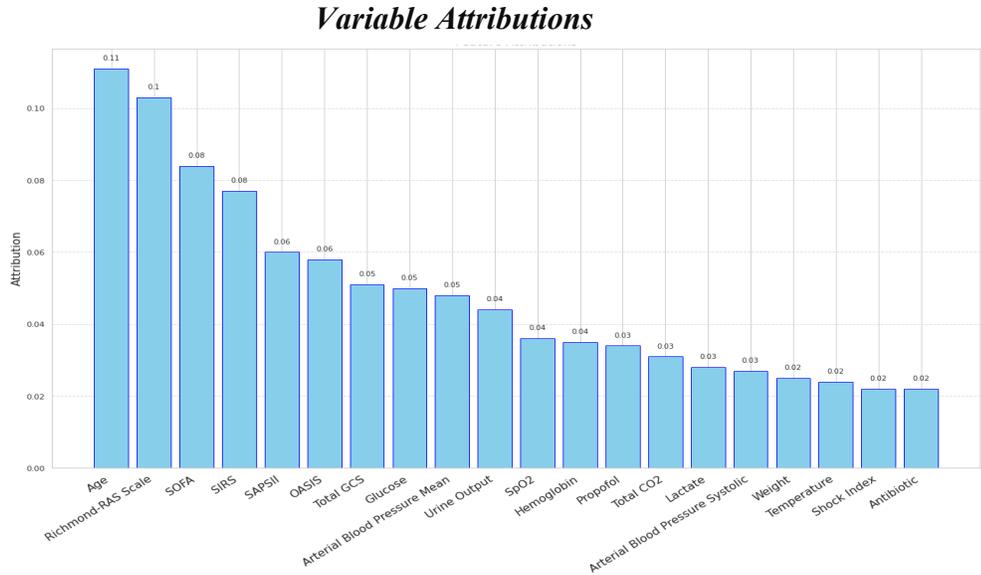

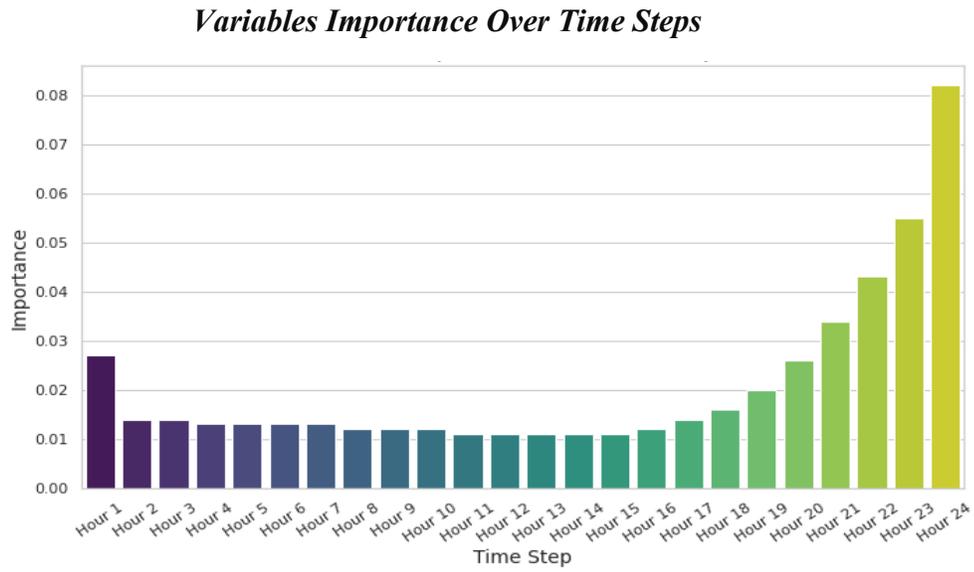

**b**      *Variable importance - external validation on eICU dataset*

*Variable Attributions*

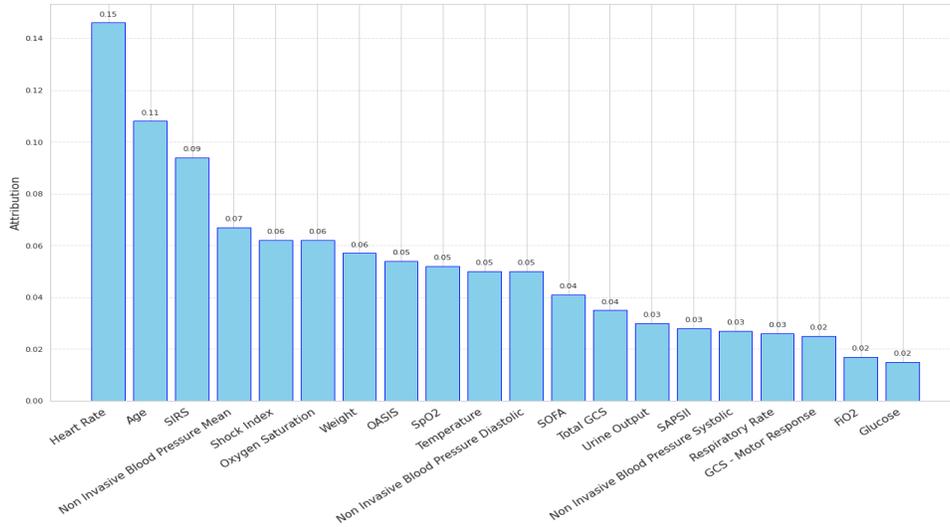

*Variables Importance Over Time Steps*

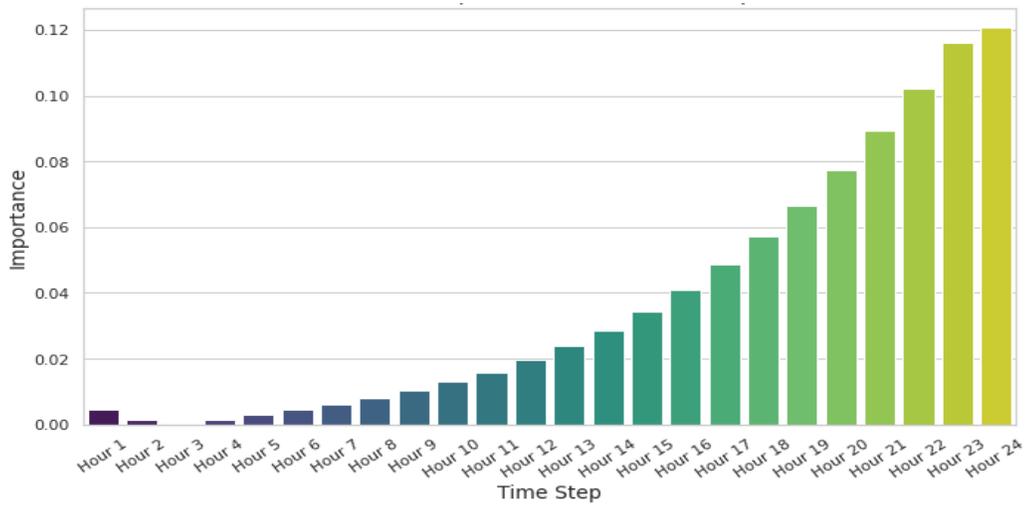

**Supplement Figure 8:** *Model performance when incorporating of clinical notes for predicting mortality.* For MIMIC-III data, a) illustrates the AUROC and AUPRC of the models when incorporating only clinical notes, only time-variant data, versus combined; b) the true negative (TNR), true positive (TPR), false negative (FNR), and false positive (FPR) rates when the model was tested on specific cohorts based on race/ethnicity, age group, or sex on data. For MIMIC-IV data, b) model performance; and d) fairness/bias as above.

## MIMIC-III

**a**

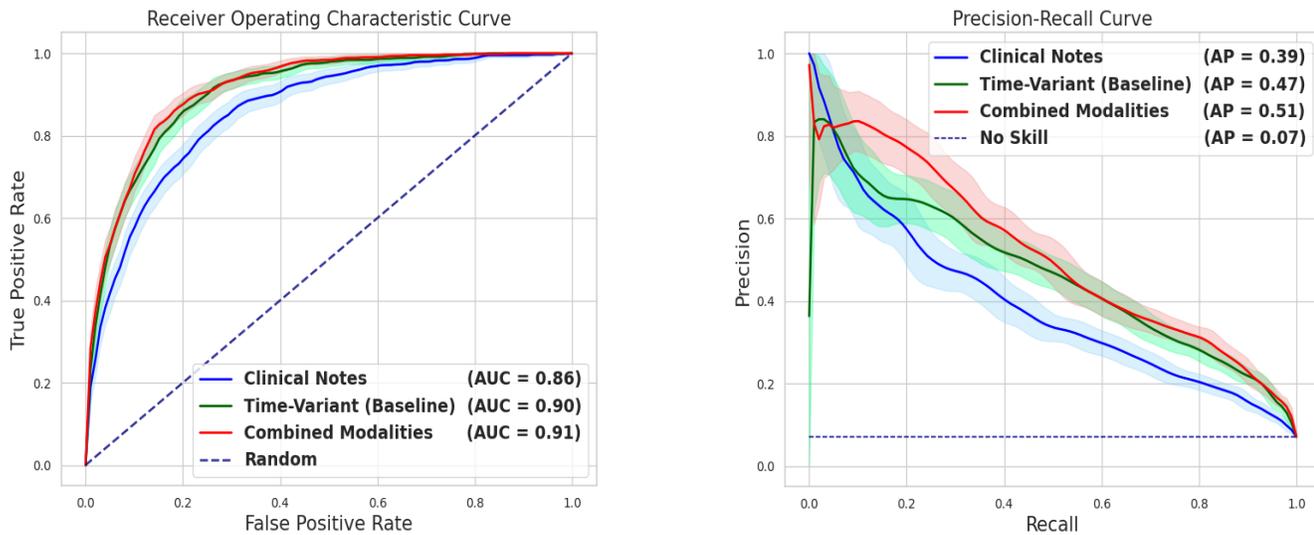

**b**

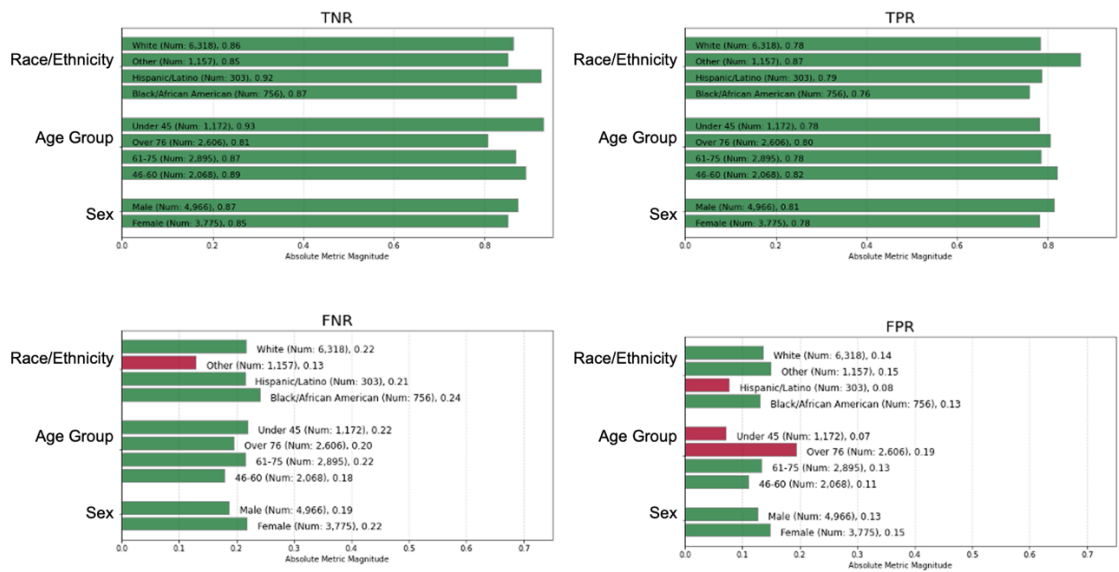

**MIMIC-IV**

**c**

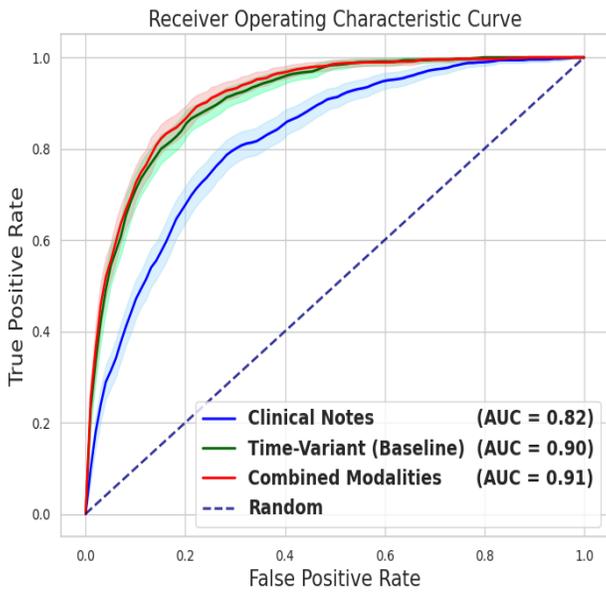
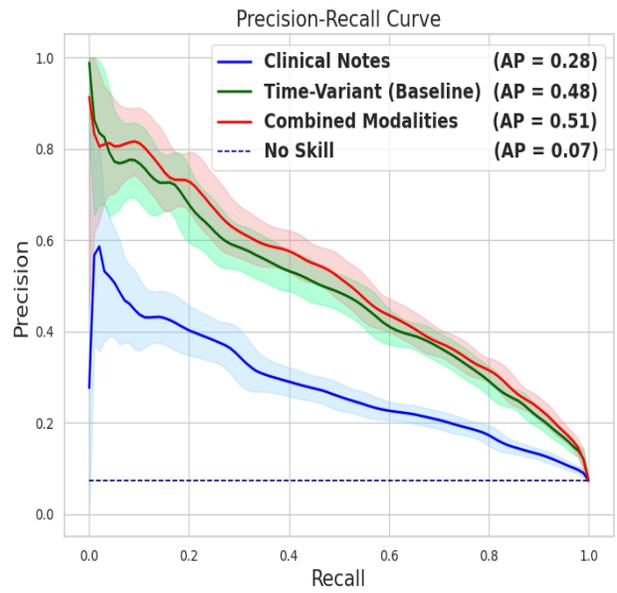

**d**

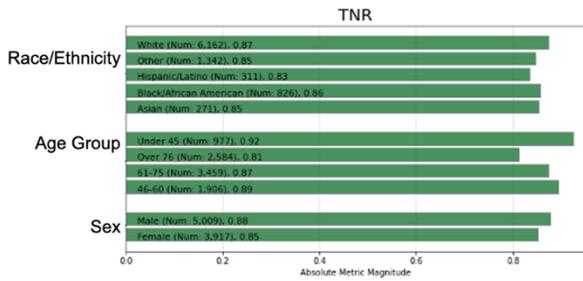
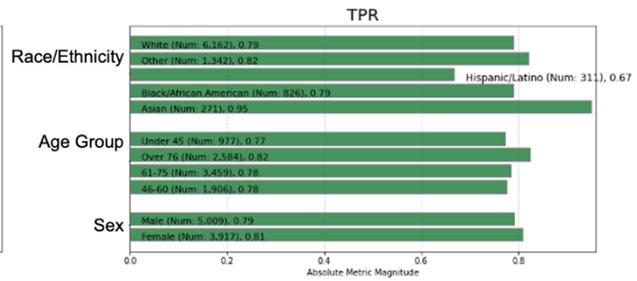

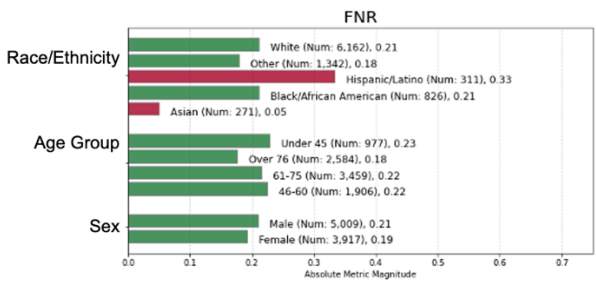
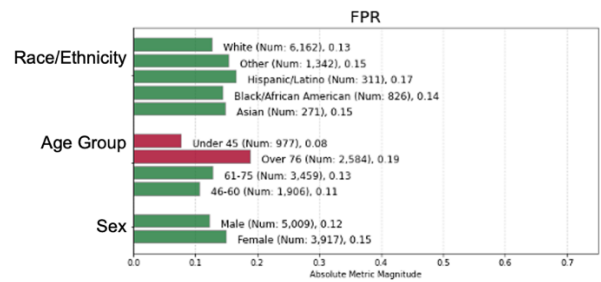

**Supplement Table 7.** *Performance metrics for incorporation of clinical notes for predicting mortality in MIMIC-III and MIMIC-IV datasets.*

**MIMIC-III:** The analyzed cohort comprised 44,074 ICU stays, including 25,098 (56.9%) male and 18,976 (43.1%) female patients. The observed mortality rate was 7%. The median age of patients was 64 years, with an interquartile range (IQR) of 53 to 78 years.

| MIMIC-III | | | | | | | | |
|---|---|---|---|---|---|---|---|---|
| Model | ACC (95% CI) | PRC (95% CI) | F-1 (95% CI) | SPE (95% CI) | AUC (95% CI) | AP (95% CI) | MCC (95% CI) | Brier (95% CI) |
| **Clinical Notes** | 0.763 (0.754-0.772) | 0.204 (0.188-0.220) | 0.325 (0.303-0.347) | 0.760 (0.751-0.770) | 0.859 (0.843-0.873) | 0.391 (0.350-0.430) | 0.321 (0.298-0.343) | 0.136 (0.130-0.141) |
| **Time-Variant Baseline** | 0.837 (0.829-0.844) | 0.278 (0.256-0.297) | 0.413 (0.386-0.436) | 0.839 (0.831-0.847) | 0.905 (0.894-0.914) | 0.468 (0.426-0.511) | 0.409 (0.385-0.433) | 0.267 (0.261-0.273) |
| **Combined Modalities** | 0.860 (0.852-0.867) | 0.313 (0.290-0.334) | 0.449 (0.424-0.473) | 0.864 (0.857-0.871) | 0.913 (0.903-0.923) | 0.512 (0.471-0.553) | 0.442 (0.418-0.467) | 0.080 (0.077-0.084) |

**MIMIC-IV:** The analyzed cohort comprised 46,765 ICU stays, including 26,892 (57.5%) male and 19,873 (42.5%) female patients. The overall mortality rate was 7%. Patients had a median age of 65 years, with an interquartile range (IQR) of 56 to 77 years.

| MIMIC-IV | | | | | | | | |
|---|---|---|---|---|---|---|---|---|
| Model | ACC (95% CI) | PRC (95% CI) | F-1 (95% CI) | SPE (95% CI) | AUC (95% CI) | AP (95% CI) | MCC (95% CI) | Brier (95% CI) |
| **Clinical Notes** | 0.710 (0.700-0.720) | 0.175 (0.161-0.189) | 0.287 (0.267-0.306) | 0.703 (0.693-0.713) | 0.819 (0.803-0.834) | 0.282 (0.249-0.315) | 0.277 (0.258-0.298) | 0.215 (0.209-0.221) |
| **Time-Variant Baseline** | 0.831 (0.824-0.839) | 0.277 (0.258-0.297) | 0.413 (0.390-0.437) | 0.833 (0.825-0.840) | 0.905 (0.895-0.915) | 0.482 (0.443-0.520) | 0.409 (0.388-0.432) | 0.307 (0.300-0.314) |
| **Combined Modalities** | 0.861 (0.854-0.868) | 0.320 (0.297-0.342) | 0.456 (0.430-0.481) | 0.866 (0.858-0.873) | 0.913 (0.903-0.923) | 0.507 (0.467-0.547) | 0.448 (0.424-0.472) | 0.058 (0.055-0.061) |

Abbreviations: CI, confidence interval; ACC, Accuracy; PRC, Precision; SPE, Specificity; AUC, Area under the Receiver Operating Characteristics Curve; AP, Area under the Precision-Recall Curve; MCC, Matthews Correlation Coefficient.

**Supplementary Figure 9:** *Explanatory models for predicting mortality using clinical note*s

A 77-year-old male patient who stayed in the ICU for 8 days had a 75% predicted risk of mortality and ultimately died in hospital.

A 71-year-old female patient who stayed in the ICU for 3 days had a 15% predicted risk of mortality and survived.

**Supplement Figure 10:** *Incorporation of medical imaging for predicting mortality using MIMIC-IV data:* a) AUROC and AUPRC of model performance; and b) fairness/bias analyses

**a**

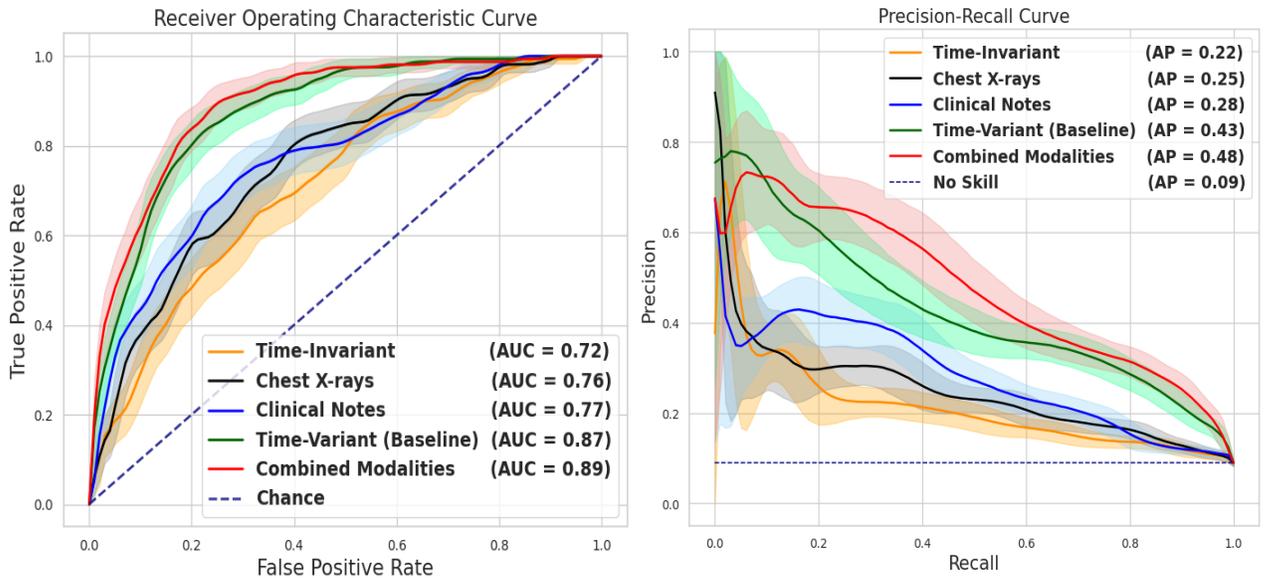

**b**

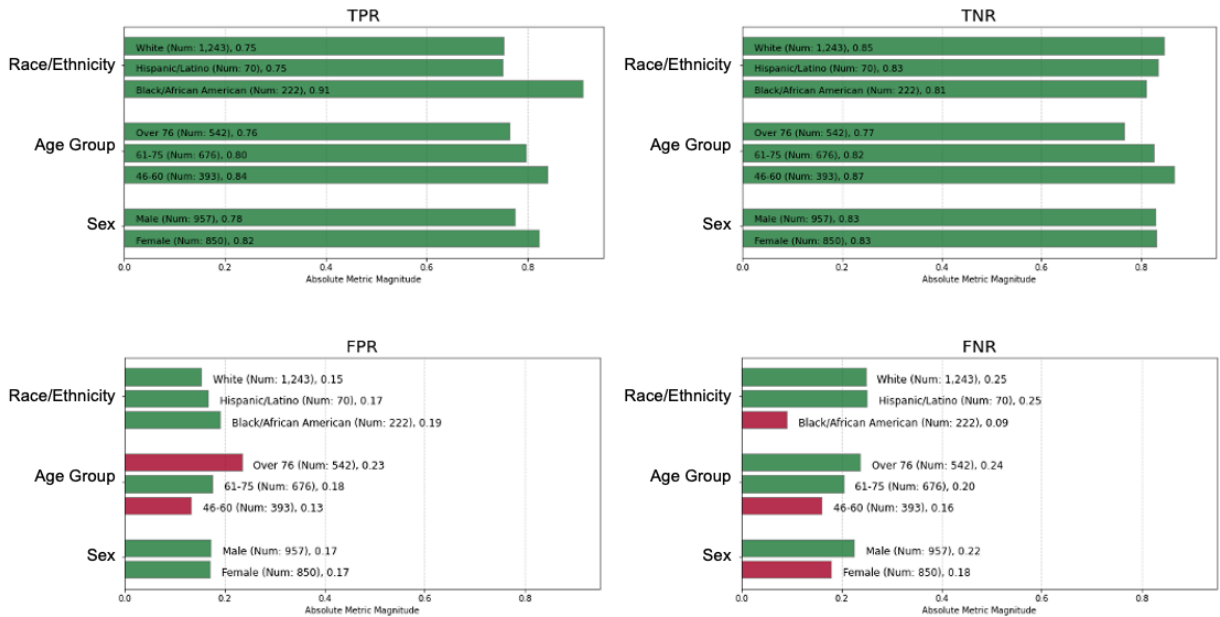

**Supplement Table 8:** *Performance metrics for incorporation of medical imaging for predicting mortality using MIMIC-IV data*

**MIMIC-IV:** The analyzed cohort comprised 9,881 ICU stays, including 5,491 (55.6%) male and 4,390 (44.4%) female patients. The mortality rate observed in this cohort was 9%. Patients had a median age of 65 years, with an interquartile range (IQR) of 55 to 78 years.

| MIMIC-IV | | | | | | | | |
|---|---|---|---|---|---|---|---|---|
| Model | ACC (95% CI) | PRC (95% CI) | F-1 (95% CI) | SPE (95% CI) | AUC (95% CI) | AP (95% CI) | MCC (95% CI) | Brier (95% CI) |
| **Time-Invariant** | 0.507 (0.484-0.530) | 0.135 (0.114-0.156) | 0.233 (0.199-0.264) | 0.474 (0.450-0.499) | 0.718 (0.681-0.755) | 0.222 (0.172-0.282) | 0.177 (0.140-0.215) | 0.404 (0.394-0.414) |
| **Chest X-ray Images** | 0.625 (0.601-0.647) | 0.167 (0.142-0.193) | 0.276 (0.240-0.313) | 0.607 (0.584-0.632) | 0.758 (0.721-0.794) | 0.252 (0.196-0.313) | 0.234 (0.193-0.273) | 0.253 (0.244-0.264) |
| **Clinical Notes** | 0.624 (0.602-0.646) | 0.165 (0.140-0.191) | 0.273 (0.237-0.310) | 0.608 (0.585-0.631) | 0.773 (0.735-0.811) | 0.288 (0.231-0.354) | 0.229 (0.186-0.269) | 0.271 (0.258-0.285) |
| **Time-Variant Baseline** | 0.789 (0.771-0.807) | 0.274 (0.237-0.311) | 0.410 (0.365-0.455) | 0.786 (0.766-0.804) | 0.874 (0.847-0.897) | 0.434 (0.358-0.506) | 0.390 (0.345-0.434) | 0.374 (0.359-0.390) |
| **Combined Modalities** | 0.827 (0.810-0.844) | 0.316 (0.271-0.360) | 0.452 (0.399-0.502) | 0.830 (0.812-0.848) | 0.891 (0.866-0.913) | 0.482 (0.398-0.555) | 0.427 (0.378-0.476) | 0.173 (0.161-0.187) |

Abbreviations: CI, confidence interval; ACC, Accuracy; PRC, Precision; SPE, Specificity; AUC, Area under the Receiver Operating Characteristics Curve; AP, Area under the Precision-Recall Curve; MCC, Matthews Correlation Coefficient.

**Supplementary Figure 11:** *Explanatory models for predicting mortality using chest X-rays*

An 80-year-old female patient who stayed in the ICU for 2 days had an 80% predicted risk of mortality and died.

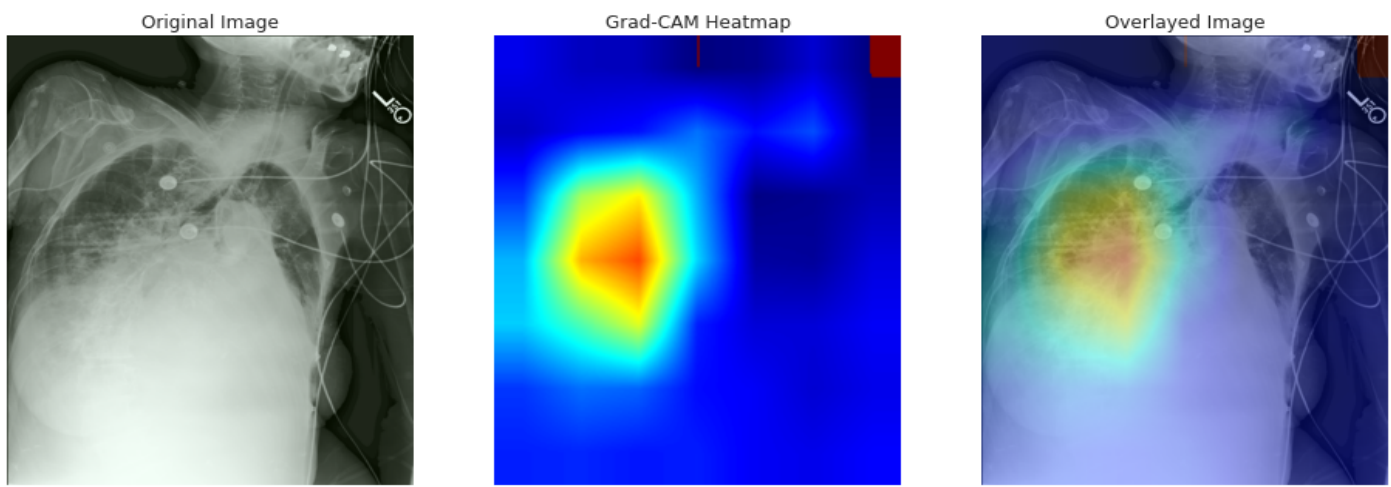

**Supplement Figure 12:** *Comparison of deep learning model to current validated risk scores*

Furthermore, we focused on the cases in which there was disagreement between our deep learning model and prediction from SAPS II (selected as its risk score outperformed OASIS and SOFA) in both MIMIC datasets. In this analysis, there were 7,348 cases where there was disagreement in ICU outcome prediction. Using these cases, we calculated the AUROC and AUPRC for each of the two models for predicting mortality, in which our model outperformed SAPS II in AUROC (0.84 versus 0.31) and AUPRC (0.26 versus 0.03) (**Figure a**). Next, we computed the absolute differences in predicted probabilities for each case between SAPS II and our deep learning model. We then focused on the cases with the largest differences (top 20% percentile). This comprised 4,138 ICU stays. Our model again outperformed SAPS II in this sub-analysis with AUROC (0.91 versus 0.27) and AUPRC (0.36 versus 0.02). (**Figure b**). Finally, we focused on the top 10% of the cases in our model that predicted the highest probability as well as the top 10% of cases in which our model predicted the lowest probability. In this sub-analysis, our model outperformed SAPS II in both AUROC (0.88 versus 0.51) and AUPRC (0.57 versus 0.18) (**Figure c**).

**a**

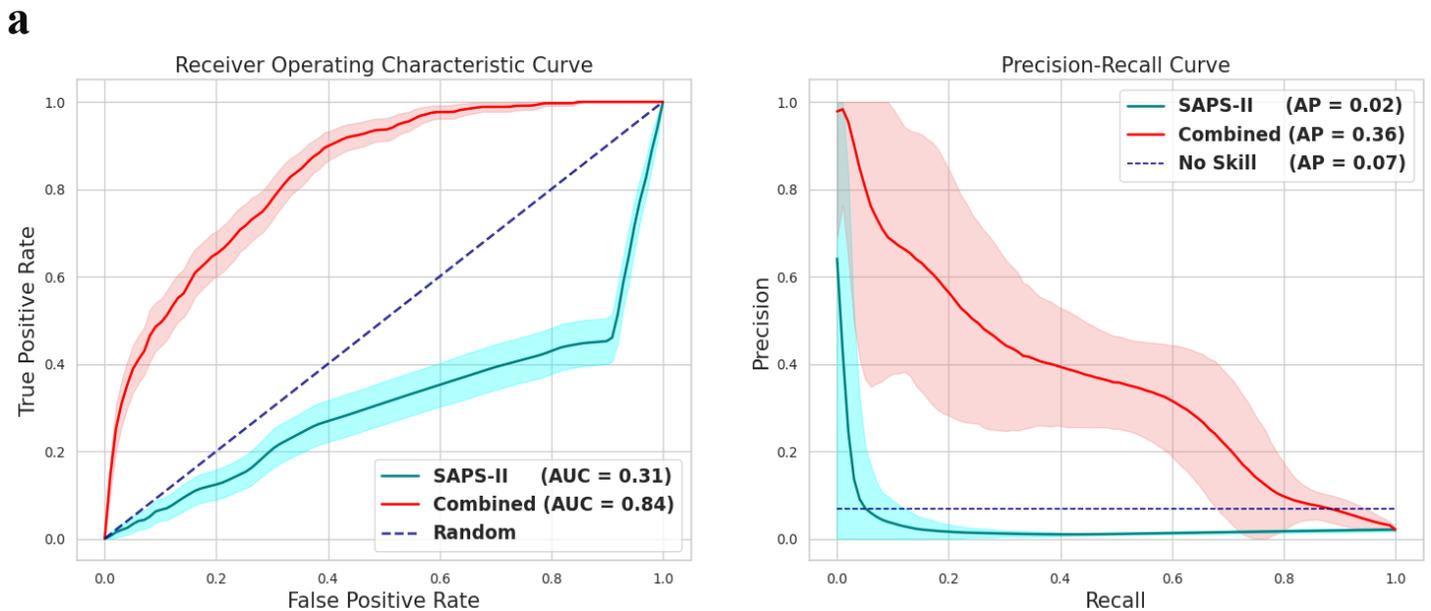

**b**

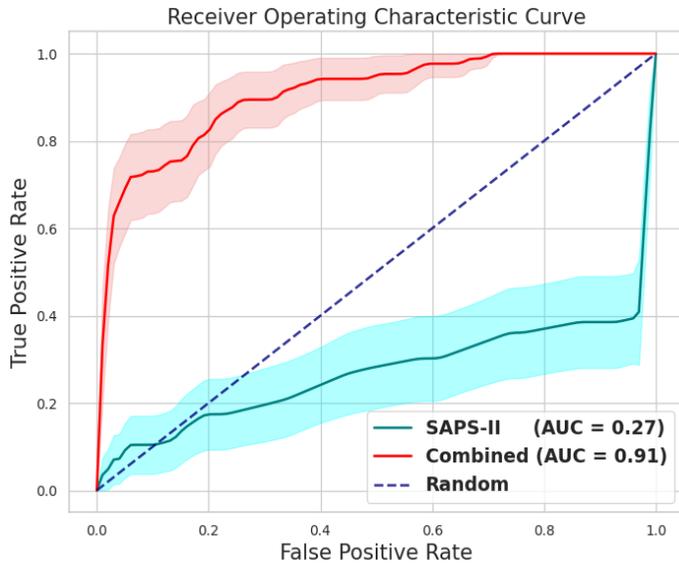

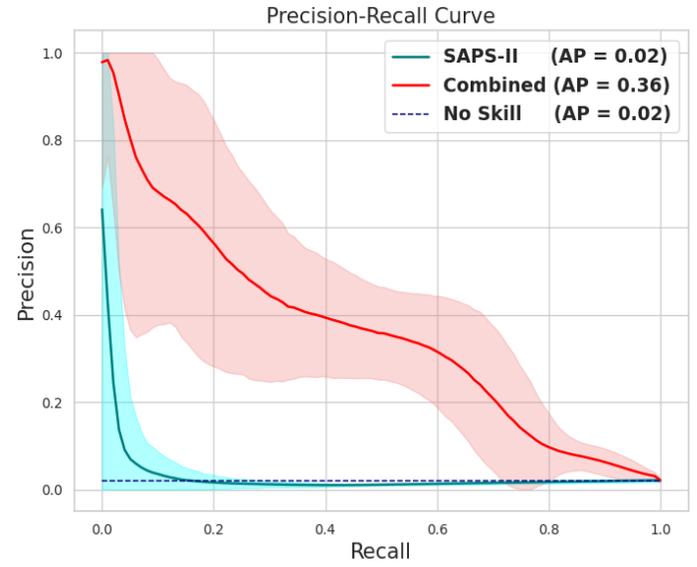

**c**

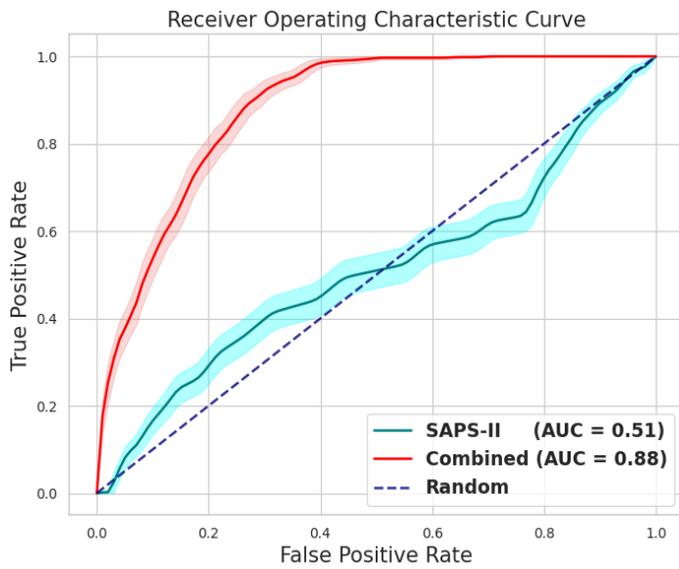

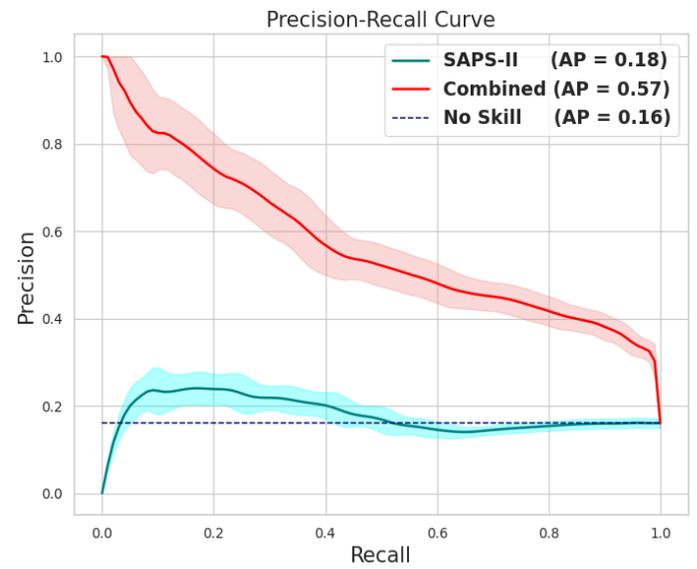

**Supplement Text 3**. *Discussion on data harmonization*

One of the primary challenges in external validation is the heterogeneity of real-world datasets, which could differ significantly from the training data. Data harmonization is essential for externally validating predictive models that integrate time dynamic data, especially hemodynamic data. It ensures consistency across disparate datasets and enhances the reliability of performance assessments. Harmonization involved extensive data pre-processing to account for inter-database differences in recorded frequency of time dynamic features (e.g., vital signs, medications), unit measurements, and rate of missingness. Without harmonization, variations in data representations can introduce biases that may subsequently obscure true generalizability. By standardizing feature inputs and aligning data distributions, we improved model comparability and facilitated a more trustworthy assessment of model robustness in external data. Furthermore, harmonization helps identify potential weaknesses in the model by ensuring that validation results are reflective of real-world variability rather than inconsistencies in data structure. Importantly, harmonization enhances the reproducibility of our multimodal deep learning model by providing a systematic approach to reconciling differences from multiple external datasets.

Given that institutions may inevitably have discrepancies in electronic health record data representation, we prioritized steps in data harmonization to enable our external validation. We developed a pipeline for quality data harmonization between distinct datasets. Harmonization of data allowed us to externally validate the models trained on MIMIC databases using HiRID, and eICU. Our findings suggest that a multimodal approach including static data and temporally dynamic structured data with appropriate data processing may lead to more accurate and robust mortality prediction with minimal algorithmic bias.

In addition to generalizability, we addressed other key ethical points regarding AI in our study design. Poorly trained AI models may frequently display algorithmic bias, which may lead to predictions that may not be equally accurate across different demographic groups, such as sex, race, and age. This bias can perpetuate disparities, which may undermine trust and efficacy in diverse patient populations. In addition, AI models are often perceived by clinicians as 'black boxes', lacking transparency and interpretability, which complicates their integration into clinical decision-making processes. Consequently, significant enhancements in the generalizability, fairness, accuracy, and interpretability of AI models, along with a reduction in algorithmic bias, are essential for their acceptance as valuable tools in clinical care. Thus, our goal was to develop a trustworthy multimodal deep learning model for predicting mortality among the critically ill by addressing some of these ethical points. As discussed previously, we first prioritized performing critical steps for data harmonization. Our harmonization pipeline addressed discrepancies across four major ICU datasets, focusing on a vital-sign frequency normalization pipeline adapted to high-resolution ICU data, medication name reconciliation, outlier removal, data upsampling/downsampling, and handling of missing values via clinically informed imputation. Then, we demonstrated generalizability by externally validating the models at different institutional data. We were transparent with algorithmic bias towards groups based on race, ethnicity, and sex. Finally, we demonstrated a pipeline for explainability of feature importance for structured data, time, clinical notes, and CXR imaging. These rigorous and detailed steps are necessary to advance AI-based clinical decision support tools into clinical practice.